\documentclass{article}

\usepackage{microtype}
\usepackage{graphicx}
\usepackage{booktabs} % for professional tables

\usepackage{hyperref}

\usepackage[accepted]{icml2024}

\usepackage{amsmath}
\usepackage{amssymb}
\usepackage{mathtools}
\usepackage{amsthm}

\usepackage{amsfonts}
\usepackage{algorithm}
\usepackage{algorithmic}
\usepackage{xcolor}
\usepackage{bm}
\usepackage{bbm}
\usepackage{diagbox}
\usepackage{wrapfig}
\usepackage{soul}
\usepackage{comment}
\usepackage{subcaption}
\usepackage{array,booktabs}
\usepackage{multicol}
\newcolumntype{M}[1]{>{\centering\arraybackslash}m{#1}} % w/ horizontal centering

\DeclareMathOperator*{\argmax}{arg\,max}
\DeclareMathOperator*{\argmin}{arg\,min}
\DeclareMathOperator{\arctantwo}{arctan2}

\definecolor{mitred}{rgb}{0.64, 0.12, 0.20}

\usepackage[capitalize,noabbrev]{cleveref}

\theoremstyle{plain}

\theoremstyle{definition}

\theoremstyle{remark}

\usepackage[textsize=tiny]{todonotes}

\icmltitlerunning{Boundary Exploration for Bayesian Optimization With Unknown Physical Constraints}

\begin{document}

\twocolumn[
\icmltitle{Boundary Exploration for Bayesian Optimization With Unknown Physical Constraints}



\icmlsetsymbol{equal}{*}

\begin{icmlauthorlist}
\icmlauthor{Yunsheng Tian}{mit}
\icmlauthor{Ane Zuniga}{mit}
\icmlauthor{Xinwei Zhang}{ibm}
\icmlauthor{Johannes P. Dürholt}{evonik}
\icmlauthor{Payel Das}{ibm}
\icmlauthor{Jie Chen}{ibm}
\icmlauthor{Wojciech Matusik}{mit}
\icmlauthor{Mina Konaković Luković}{mit}
\end{icmlauthorlist}

\icmlaffiliation{mit}{MIT CSAIL, USA}
% \icmlaffiliation{umn}{University of Minnesota, USA}
\icmlaffiliation{evonik}{Evonik Operations GmbH, Germany}
\icmlaffiliation{ibm}{MIT-IBM Watson AI Lab, IBM Research, USA}

\icmlcorrespondingauthor{Yunsheng Tian}{yunsheng@csail.mit.edu}
\icmlcorrespondingauthor{Mina Konaković Luković}{minakl@mit.edu}

\icmlkeywords{Machine Learning, ICML}

\vskip 0.3in
]



\printAffiliationsAndNotice{}  % leave blank if no need to mention equal contribution

\begin{abstract}
Bayesian optimization has been successfully applied to optimize black-box functions where the number of evaluations is severely limited. However, in many real-world applications, it is hard or impossible to know in advance which designs are feasible due to some physical or system limitations. These issues lead to an even more challenging problem of optimizing an unknown function with unknown constraints. In this paper, we observe that in such scenarios optimal solution typically lies on the boundary between feasible and infeasible regions of the design space, making it considerably more difficult than that with interior optima. Inspired by this observation, we propose \emph{BE-CBO}, a new Bayesian optimization method that efficiently explores the boundary between feasible and infeasible designs. To identify the boundary, we learn the constraints with an ensemble of neural networks that outperform the standard Gaussian Processes for capturing complex boundaries. Our method demonstrates superior performance against state-of-the-art methods through comprehensive experiments on synthetic and real-world benchmarks. Code available at: \url{https://github.com/yunshengtian/BE-CBO}
\end{abstract}

\section{Introduction}

Many optimization problems involve the need to optimize black-box functions, where the performance of a sample can only be determined through physical experimentation or time-expensive simulation, which may take even on the order of weeks or months per single experiment. Thus, the total number of evaluations that can be conducted is limited. In this scenario, Bayesian optimization (BO) \citep{jones1998efficient,BOsurvey} has proven to be a successful approach that guides the search for an optimal solution by iteratively proposing which experiment to evaluate that may lead to the highest performance increase.

In addition to optimizing unknown functions, many practical problems include unknown constraints. For untested samples, it is impossible to determine if a particular combination of design parameters will lead to feasible or infeasible design. In such cases, infeasible regions are typically discovered when pushing the limits of what is physically possible, and the optimal solution lies on the boundary of feasible regions. 

For example, consider the problem of designing an airplane wing. The goal is to make it as lightweight as possible while also being durable and structurally strong to sustain the forces. Hence, the optimization needs to reduce the amount of used material until the wing starts breaking under applied forces. The optimal design will be found just before we step into the infeasible range. Similarly, imagine the task of moving a linear stage to a desired position by applying an acceleration followed by deceleration. To optimize this process, the optimizer might progressively increase the voltage applied to achieve faster and more efficient movement. However, if the voltage is continuously increased without considering the motor's limitations, it can lead to excessive heat generation, effectively burning the motor and causing performance degradation or failure. Comparable examples can be found in chemistry and materials science, especially in formulation development where often certain amounts/combinations of ingredients are needed to yield feasible materials, and the actual properties of interest can be measured only for feasible materials. This leads to a situation in which the optimizer is proposing a candidate for which no performance feedback can be obtained, resulting in a BO iteration in which no new information is retrieved. 
Indeed, while analyzing many real-world benchmark problems, we observe that all these problems have the optimal solution on the boundary between feasible and infeasible designs (further details in Section~\ref{sec:discussion}).

Motivated by this observation, we argue that a good model of the boundary and efficient search around it can significantly improve efficacy and lead to the discovery of better solutions compared to exploring the entire design space. We propose a novel BO algorithm incorporating a boundary exploration strategy tailored for exploiting this problem structure, which is prevalent in real-world problems with unknown physical constraints. 
In case of unknown physical constraints we need to model the space of feasible and infeasible designs. In physical experiments a sample is considered infeasible if it fails before being able to measure its performance, therefore such sample does not provide any continuous value in return. Hence, we train a binary classifier to represent all constraints that may appear in the system. Furthermore, it is important to note the difference between imposed constraints and scenarios where the constraints are unknown. In the former, when the constraints are specified by the user, imposed artificially, the optimum solution may be found in the interior of the feasible region. The case of boundary optima is considerably more challenging than that of interior optima, more even so for BO, where gradients are infeasible to evaluate and the number of function  evaluations is subject to a limited budget. For unknown constraints, a surrogate needs to be sufficiently accurate (at least in the vicinity of the optimum) so that the identified solution is sufficiently close to the feasible side of the boundary while not trespassing on the other side. 

While existing approaches \citep{gelbart2014bayesian,2021Scalable,antonio2021sequential} have addressed BO with unknown constraints, our method is the first to explicitly consider the boundary issue and it demonstrates superior performance. 
In summary, our main contributions are the following:
\begin{itemize}
    \item We introduce BE-CBO (Boundary Exploration for Constrained Bayesian Optimization) for optimizing black-box functions with a limited evaluation budget and unknown constraints. Our key insight is that accurately modeling and efficiently exploring the boundary between feasible and infeasible designs is crucial in discovering better performing designs. 
    \item To model the unknown constraint boundary, we propose using \emph{Deep Ensembles} and demonstrate its superior modeling capability by comparing against the most common method in surrogate modeling for BO, i.e., Gaussian Processes.
    \item Comprehensive experiments and ablation studies on synthetic functions and real-world benchmark problems showing the efficiency of our method and state-of-the-art performance on practical setups.
\end{itemize}

\section{Related Work}
\label{sec:related_work}

\paragraph{Constrained Bayesian Optimization}
BO has proven to be a powerful methodology for global optimization of black-box functions with expensive evaluations~\citep{BOsurvey}. It has demonstrated remarkable success in various applications, including robotics~\citep{Lizotte2007}, resource allocation~\citep{wrro151331}, hyperparameter tuning~\citep{snoek2012practical}, experimental design~\citep{srinivas2010gaussian}, and clinical drug trials~\citep{yu2019drugs}. 
In classical BO, the feasible set $\mathcal{X}$ is assumed to be known and easy to evaluate, e.g., they are either a hyper-cubes or a simplex~\citep{movckus1975bayesian,frazier2018bayesian}.
Recent research faces a more challenging scenario called constrained Bayesian optimization (CBO), where the feasiblility of optimization variables is unknown or hard to evaluate. In these cases, the evaluation of the feasibility of one solution is also time costly as the objective function. Two slightly different settings are considered in CBO: 1) continuous-valued constraint and 2) binary-valued constraint.
In continuous-valued CBO, the constraints take the form of inequality constraint $R(x)\geq 0,$ where $R:\mathbb{R}^d\rightarrow \mathbb{R}$ is a continuous-valued function. It considers the scenario where both the function value and the constraints' values can be obtained from the experiments, even if the sample falls into the infeasible region. In the binary-constraint case, the outcome of the experiment becomes $\mathbf{1}_{R(x)\geq 0}f(x)$, that we only observe the function value when the sample is feasible ($R(x)\geq 0$), and when the sample falls into the infeasible region, we observe a failure with no function value. 

Prior works mainly consider continuous-valued constraints. \citet{gardner2014bayesian} proposes EIC that multiples the Expected Improvement (EI) \citep{movckus1975bayesian} with the probability of constraint satisfaction as the acquisition function. PESC~\citep{hernandez2015predictive} extends Predicted Entropy Search (PES)~\citep{hernandez2014predictive} to the constrained case. ADMMBO~\citep{ariafar2019admmbo} uses ADMM to alternatively optimize the objective value and feasibility of the solution. SCBO~\citep{2021Scalable} uses a trust region optimizer to scale up to high dimensional problems with constraints.
2-OPT-C~\citep{zhang2021constrained} applies a multi-step lookahead approach instead of the standard myopic approach, which encourages sampling the boundary between feasible and infeasible regions, sharing a similar motivation as our method.
All methods use Gaussian Process (GP) regressors to model the unknown constraints.

The binary-valued CBO is much less considered in the prior works, though the setup is prevalent in practice for optimizing physical systems. Modeling the binary-valued constraint of the feasibility set is fundamentally different from modeling the continuous-valued constraints. 
Similar to \citet{gardner2014bayesian}, \citet{gelbart2014bayesian} proposes CEI that multiples the EI with the probability of constraint satisfaction.
\citet{lindberg2015optimization} uses a slightly different formulation by using asymmetric entropy as the feasibility score for efficient exploration. However, these methods use GP classifiers (GPC) for relatively simple and low dimensional constraints and the efficacy on complicated constraints is not demonstrated. A recent work SVM-CBO~\citep{antonio2021sequential} proposes a two-stage approach. In the first stage, it trains an SVM classifier to explore the constraint boundary, and in the second stage, the algorithm performs BO in the feasible region captured by the SVM.

\paragraph{Safe Bayesian Optimization} In some scenarios, ensuring the safety and reliability of evaluations have become important concerns when running BO. To address these concerns, safe BO techniques have emerged as a promising solution~\citep{sui2015safe, turchetta2019safe, kirschner2019adaptive}. A representative application of safe BO can be found in robotics~\citep{baumann2021gosafe, berkenkamp2023bayesian} due to the concern of robot hardware damage during optimization. However, although constrained BO and safe BO share a similar spirit, the main objective in constrained BO is to achieve higher performance while safe BO puts more priority on being safe, i.e., conservative on violating the constraint and exploring infeasible design space, which will be less effective in terms of finding the optimum. 

\paragraph{Neural Networks for Classification}
Though GPC has been used in CBO algorithms for modeling the constraints~\citep{bachoc2020gaussian,lindberg2015optimization}, they face the issue of high computational complexity in both theory and lack of supporting software packages, and require a careful choice of the GP kernel, or otherwise the resulting constraint boundaries might be too smooth and lack details. In contrast, Neural Networks (NN) have been widely used and have reached great success in classification problems with both low- and high-dimensional data with complex boundaries, including manually selected features~\citep{swain2012approach} and raw data such as images~\citep{lu2007survey}, texts~\citep{minaee2021deep}, and graphs~\citep{zhang2018end}. Despite impressive classification accuracies in supervised learning benchmarks, naive NNs are poor at quantifying predictive uncertainty~\citep{lakshminarayanan2017simple,gawlikowski2023survey}, thus cannot be directly applied to CBO for constraint modeling. Recently, the prediction uncertainty of NNs has been studied in several approaches, including Bayesian NN~\citep{kwon2020uncertainty,tran2019bayesian} and ensemble methods~\citep{lakshminarayanan2017simple}, have been proposed. Additionally, the ensemble method has shown strong promise in few-shot training~\citep{beluch2018power} where the number of training samples is small.

\section{Preliminaries}
\label{prelim}

We are interested in efficiently optimizing black-box functions with costly evaluations and unknown constraints. Bayesian Optimization (BO) is a powerful framework for such scenario~\citep{jones1998efficient,BOsurvey}. In the typical setup, BO addresses the challenge of finding the global optimum of an expensive objective function $f : \mathcal{X} \subset \mathbb{R}^d \rightarrow \mathbb{R}$, where direct gradient information is unavailable. 
In addition, the total number of performed function evaluations is often limited to several dozens. 

The optimization process begins with an initial set of $N$ evaluations $Y_0 = \{f(x_i)\}_{i=1}^N$, where $x_i$ is sampled at random from the design space $\mathcal{X}$, to explore the function's behavior. A surrogate model $\hat{f}(x\vert Y_0)$, typically a Gaussian Process (GP), is then used to approximate the unknown function $f$ using existing observations $Y_0$. The main strength of BO lies in the strategy for selecting the subsequent evaluations by balancing exploration and exploitation. This balancing is defined with an acquisition function $q$ that guides the search by trading off between exploiting promising regions and exploring uncertain regions of the function. Popular acquisition functions include Expected Improvement (EI)~\citep{movckus1975bayesian}, Entropy Search (ES)~\citep{hennig2012entropy}, Predictive Entropy Search (PES)~\citep{hernandez2014predictive}, and Thompson Sampling (TS)~\citep{thompson1933likelihood}. In this work, we use EI, but the proposed method is straightforward to generalize to other acquisition functions. EI measures the expected amount of improvement over the current best value $f'$ by observing at the sampling point: $\mathrm{EI}(x) = \mathbb{E}[\max\{0, \hat{f}(x\vert Y_i)-f'\}].$
When $\hat{f}(x)$ is GP, EI has a closed-form expression~\citep{jones1998efficient}, thus widely used in practice. Finally, the sample for next-step evaluation is selected as $x^+ = \mathrm{arg}\max_{x} \mathrm{EI}(x)$.
The set of observed samples is updated as $Y_{i+1} = Y_{i} \cup \{f(x^+)\}$, and $\hat{f}$ is recomputed on $Y_{i+1}$.

Another layer of complexity is added to our problem by incorporating unknown constraints. In this work, we are interested in practical applications where design samples can be feasible and infeasible. Unlike other works with constraints that are continuous functions giving a real value even for samples that do not satisfy the constraints~\citep{2021Scalable}, in our setup, it is impossible to obtain any objective value when the design is infeasible. These designs are impossible to create and evaluate. They either contradict the forces of physics or fail during the evaluation time. Hence, we model the constraints as a binary function $c: \mathcal{X} \rightarrow \{0,1\}$, where $x$ is feasible if $c(x)=1$ and infeasible otherwise. Our final goal can be formalized as 
\[\argmin_{x \in \mathcal{X}} f(x) \quad \text{s.t.} \quad c(x) = 1,\]
where both $f$ and $c$ are unknown.
\section{Proposed Method}
Our proposed method implements the BO pipeline with a few modifications described below. In summary, our method consists of the following steps: We first conduct initial evaluations on a random set of samples. We then fit a GP (as outlined in \citet{GP}) on the evaluated data to model the objective function. In parallel, we fit another surrogate model to approximate the constraints described in Section~\ref{classifier}. To select which sample to evaluate next, we optimize an acquisition function $\mathrm{EI}$ (see Section~\ref{prelim}) with a constrained optimization approach introduced in Section~\ref{sec:be}. Finally, we evaluate the proposed sample and iterate until we reach the budget for the number of function evaluations.

\subsection{Modeling Constraints}
\label{classifier}

\begin{figure}[t]
\setlength\tabcolsep{3pt} % default: 6pt
\centering
\begin{tabular}{@{} r M{0.8\linewidth} @{}}
\begin{subfigure}{0.05\linewidth} \caption{} \end{subfigure} 
  & \includegraphics[width=\hsize]{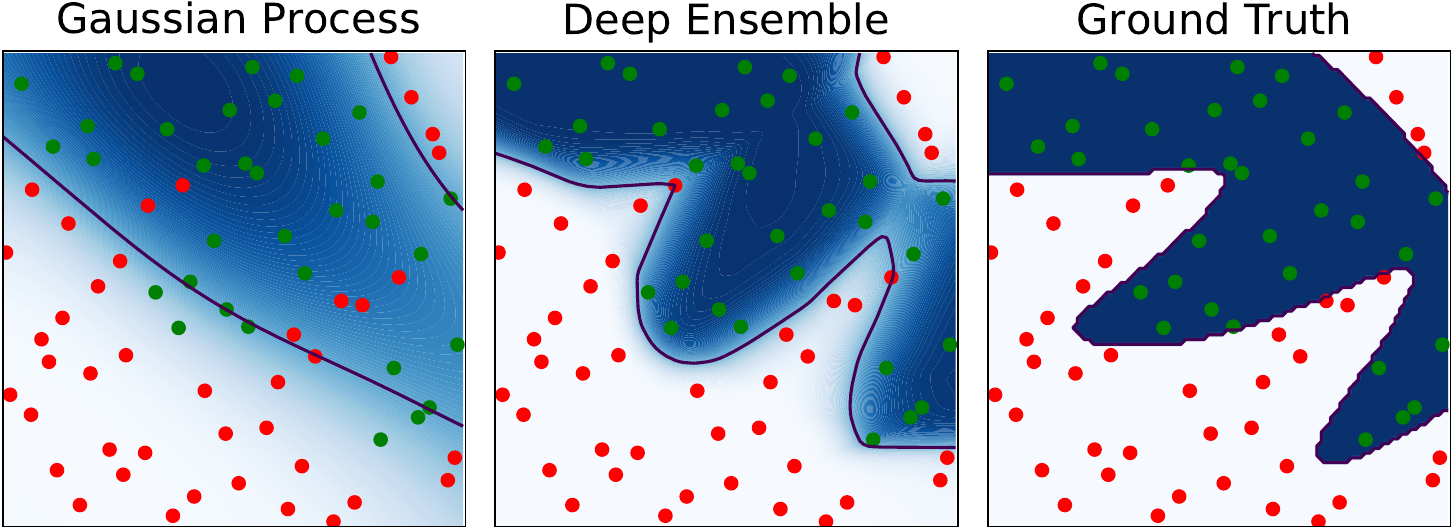}\\
\begin{subfigure}{0.05\linewidth} \caption{} \end{subfigure} 
  & \includegraphics[width=\hsize]{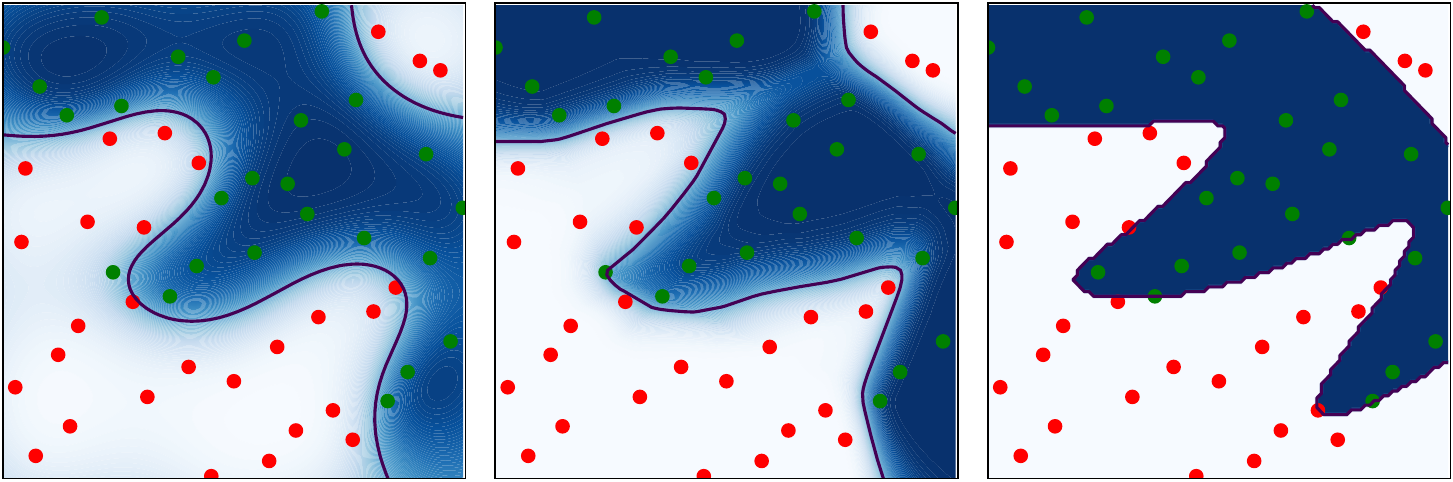}\\
\begin{subfigure}{0.05\linewidth} \caption{} \end{subfigure} 
  & \includegraphics[width=\hsize]{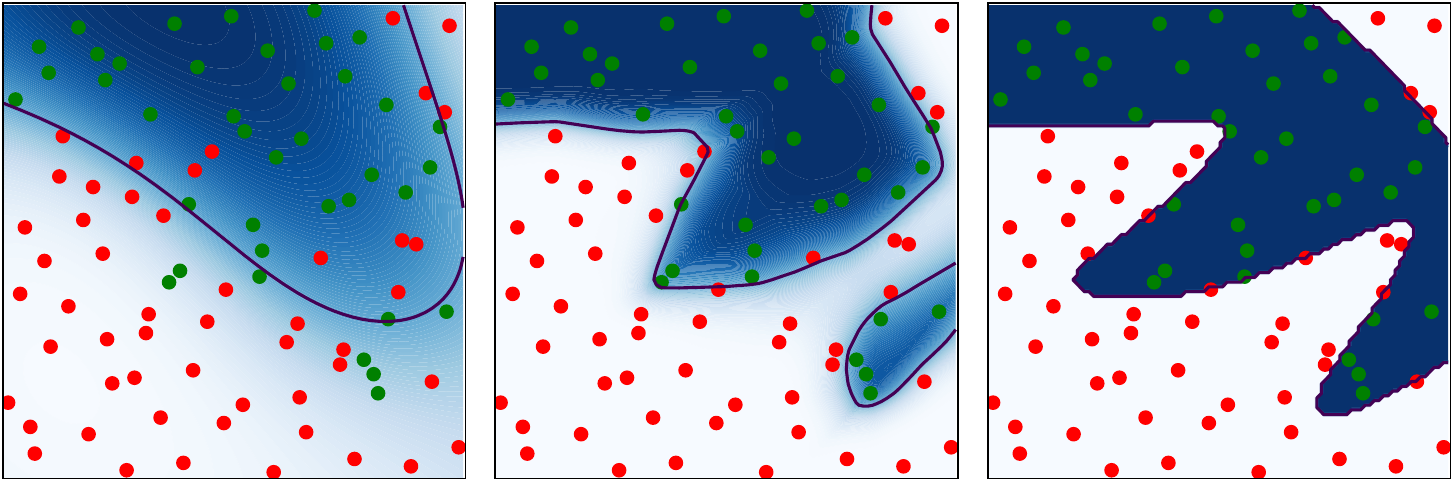}\\
\end{tabular}
\caption{Comparison of modeling constraints between Gaussian Processes (GP) and Deep Ensembles (DE) on the LSQ problem with three sets of random sample evaluations. Green dots represent feasible designs and red dots represent infeasible designs. We observe that DE tends to be more robust than GP in capturing complex boundaries. (a) Failure case of GP; (b) Successful case of GP; (c) Both GP and DE are not fitted well due to the poor sample distribution but DE is closer to the ground truth.}
\label{fig:classif}
\end{figure}

As described in Section~\ref{prelim}, we consider the problem of representing constraints as a binary function, i.e., a classifier that predicts for each design sample whether it is feasible or infeasible. Specifically, we aim to train a classifier $C: \mathcal{X} \subset \mathbb{R}^d \rightarrow [0,1]$ that measures a probability of sample $x$ being feasible. When $C(x) > 0.5$, sample $x$ is considered more likely to be feasible.

For many practical problems, it is difficult even to find a single feasible solution due to the non-convex nature of the feasible set. Hence, fitting a good surrogate model for the unknown constraints is a delicate task. Many previous works utilize GPs to approximate the constraints based on a given set of evaluations~\citep{2021Scalable,gelbart2014bayesian}. However, since we observe that the optimal solution typically lies on the boundary between feasible and infeasible regions (see Section~\ref{sec:be}), having a higher-accuracy surrogate model is crucial. To the best of our knowledge, we are the first to propose the use of \textit{Deep Ensembles (DE)}~\citep{lakshminarayanan2017simple,fort2019deep} for this purpose in Bayesian optimization. Please note the difference between ensembles previously used in Bayesian optimization that were in the context of Gaussian Process Ensembles~\citep{wang2018batched} and Acquisition Functions Ensembles~\citep{hoffman2011portfolio, kandasamy2020tuning}.

\paragraph{Deep Ensembles}
DE is composed of a set of neural networks, often multilayer perceptrons (MLPs), and can be used to model epistemic uncertainty. By training multiple independent MLPs and leveraging the diversity among them, DE enable improved accuracy and enhanced uncertainty estimation. Each MLP in the ensemble is trained on the same dataset but with different random initialization for weights. During inference, predictions from the ensemble are obtained by averaging or combining the individual model predictions. Unlike Bayesian NN, it does not require delicate hyperparameter tuning and long training.  

\begin{figure*}[t]
    \centering
    \includegraphics[width=0.9\textwidth]{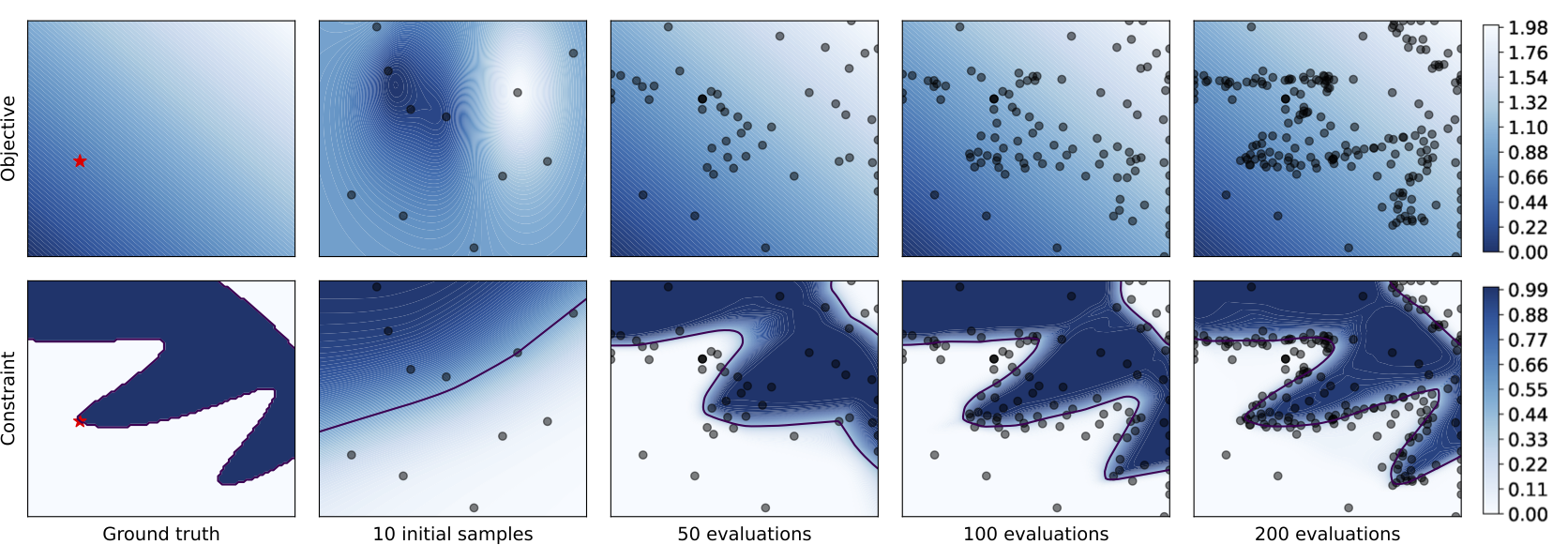}
    \caption{Example evaluation of the LSQ synthetic benchmark problem guided by our algorithm BE-CBO. Top row illustrates the objective function, while the bottom row illustrates the classifier. In the left plot, the ground truth is shown with the boundary and optimal solution. In the second plot, the state of the surrogate is shown for both the constraint and the objective, based on the first initial 10 samples. The following three plots demonstrate the state of the surrogate models for the objective and the constraint after 50, 120, and 200 evaluations.}
    \label{fig:progress}
\end{figure*}

More specifically, in this work, we train $N$ identical MLPs $M_{\theta_i}$ independently on all evaluated samples $Y_i$ to model the probability of each sample being feasible or infeasible. The output of each $M_{\theta_i}$ is a probability value between 0 and 1 at a given sample point $x$. Our classifier $C$ is defined as a DE and takes the mean of the predicted value from $M_{\theta_i}$, $\mu_E(x) = \frac{1}{N}\sum_i^N M_{\theta^\star_i}(x)$, where $\theta^\star_i$ are trained weights for each model. In addition, we can compute the variance of the prediction as $\sigma_E(x)^2 = \frac{1}{N-1}\sum_i^N (M_{\theta^\star_i}(x) - \mu_E(x))^2$. To evaluate the feasibility probability $C(x)$ we are essentially computing $C(x) = \mu_E(x)$. See details in Appendix~\ref{app:method:de}. 

% The benefits of Deep Ensembles are two-fold: fast to train due to easy parallelization and simple architectures (see Section~\ref{sec:results} for runtime comparisons). Ensembles tend to produce more accurate approximations of the classification problem at hand (see, for example, Figure~\ref{fig:classif} and refer to paper~\citep{beluch2018power} for additional examples). 
% Please see Appendix~\ref{app:ablation:gp_de} for more detailed comparison between GP and DE.

\paragraph{Benefits of Deep Ensembles}
GP-based constraint modeling has several limitations which can be addressed by using DE. First, GPs inherently assume a degree of smoothness in the underlying function they model, as evidenced by the choice of kernel function (e.g., RBF, Matern). This smoothness assumption can lead to difficulties when trying to capture sharp boundaries or discontinuities of the constraint. Second, the kernel structure is chosen a priori and remains fixed throughout the optimization. Although parameters of the kernel can be fitted to data, the form of the kernel itself is fixed and might not be well-suited to represent complex, non-linear boundaries. This limitation is discussed in depth in the context of model selection and adaptation~\cite{duvenaud2013structure}. Thus, compared to GP, DE is more flexible and powerful in representing complex constraint functions (see Figure~\ref{fig:classif} and \citep{beluch2018power} for examples). Besides, DE is also fast to train due to easy parallelization and simple architectures (see Section~\ref{sec:results} for runtime comparisons). Please refer to Appendix~\ref{app:ablation:gp_de} for more detailed comparison between GP and DE.

\paragraph{Training Deep Ensembles} 

In practice, we find that training DE with the popular maximum likelihood estimation (e.g., using binary cross-entropy loss) as suggested by \citet{lakshminarayanan2017simple} leads to poorly calibrated uncertainties and thus deteriorates the BO performance. Instead, we find training DE with variational inference, specifically, evidence lower bound (ELBO), provides better-calibrated uncertainties and improves performance by a large margin, as aligned with observations from \citet{tomczak2018neural}. This approach requires the model to output continuous latent values that can be transformed to probabilities using Bernoulli likelihood. Please see Appendix~\ref{app:method:de:mean_std} for computation details and experimental validations.

\subsection{Boundary Exploration}
\label{sec:be}
For many practical problems, there is no information about constraints, and discovering the feasible and infeasible regions of space is intertwined with the optimization process. These infeasible regions occur when the system has physical or implicit limitations not artificially imposed by the users. In such scenarios, we discover infeasible regions by pushing the limit of what is feasible as we try to optimize the designs further. This reasoning, and many examples we have seen in practice, lead us to observe that the optimum most frequently (if not always) lies on or very close to the boundary between feasible and infeasible regions.

To ensure exploration around the boundary, we formulate a constrained optimization problem:
\begin{equation}
    \argmax_x{q(x)} \quad \text{s.t.} \quad l(x) \leq C(x) \leq u(x)
    \label{const_opt}
\end{equation}
where $q$ is an acquisition function (see Section~\ref{prelim}), $C$ is the classifier (Section~\ref{classifier}), and $l$ and $u$ are lower and upper bound functions respectively. As discussed in Section~\ref{prelim}, we implement EI for the acquisition function, but note that Eq.~\ref{const_opt} is not in any way tied to the particular formulation of EI. The proposed method can easily generalize to other acquisition functions. 
Function $l$ determines how far into the infeasible region can we sample. By allowing the samples to be queried in the infeasible part it encourages pushing the boundary and discovering new regions. However, querying many infeasible samples is especially problematic for a small sampling budget as those samples do not return any value. Hence, we limit $l$ to remain close to the boundary, accounting for uncertainty in the prediction of the boundary, as discussed in the following paragraph. Function $u$, however, determines how far into the feasible region we can sample. Since all feasible values return additional information for the GP modeling the objective function, we allow exploration of the entire feasible region and set $u(x)=1$ (recall that the upper bound for $C$ is equal to $1$ since it represents the probability).
Please note that all the functions are updated in every iteration of BO to fit the new data.

A trivial approach to defining the lower bound around the boundary would be to set a constant value, say $l(x)=0.4$. This bound leaves a 10\% margin on the infeasible side of the boundary to allow some space for exploration and addressing the inaccuracy of surrogate model predictions. However, in the initial iterations of BO, the surrogate models for classifier and objective function typically exhibit very high inaccuracy due to the small number of samples for fitting and nonlinear functions to approximate (see, for example, Figure~\ref{fig:progress}). The accuracy can quickly increase with more samples being evaluated. Hence, the bound should also account for this change. It should be wider when the accuracy is low. We propose a \textit{dynamic bound} strategy, where the exploration region around the boundary directly relates to the uncertainty of the fitted classifier. Inspired by UCB~\citep{srinivas2010gaussian}, we define the dynamic bound~as:
\[l(x) = 0.5 - \sigma_E(x)\]
where $\sigma_E(x)$ is the standard deviation of the Deep Ensembles $C$ for input $x$.

\begin{figure*}[t]
    \centering
    \includegraphics[width=\textwidth]{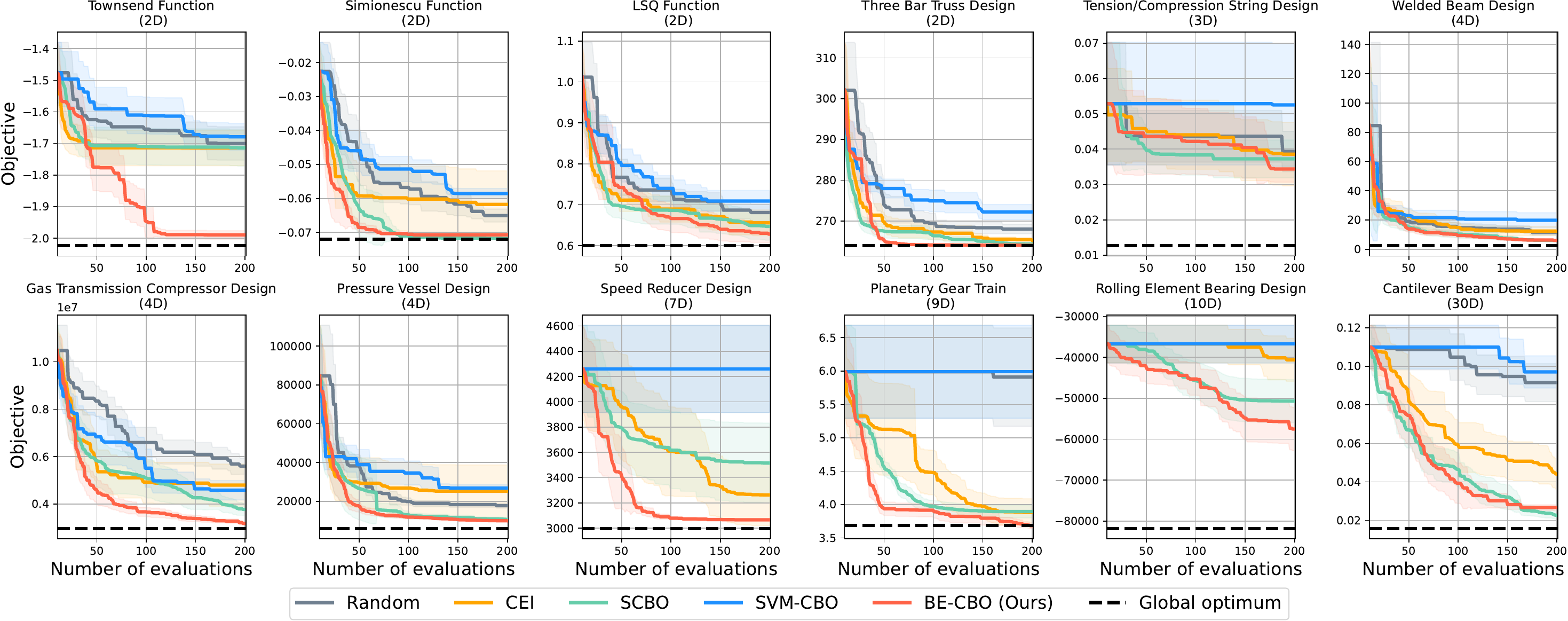}
    \caption{Quantitative comparison of different algorithms including BE-CBO on all benchmark problems. The current best value is shown w.r.t. the number of function evaluations. Each experiment has 10 initial random samples and 200 total evaluations. The curve is averaged over 10 different random seeds and the standard deviation is shown as a shaded region.}
    \label{fig:performance}
\end{figure*}

With this formulation, we incorporate the uncertainty of the classifier into the selection strategy, while the uncertainty of the surrogate of the objective function is captured in the acquisition function~\citep{movckus1975bayesian}.
We demonstrate one example of the optimization progress lead by BE-CBO over 200 evaluations in Figure~\ref{fig:progress}. It qualitatively shows that accurate constraint modeling plus active boundary exploration is effective in discovering both the correct constraint boundary and the global optimum.

\section{Experimental Evaluation}
\label{sec:results}
We conduct comprehensive experiments to evaluate the performance of our methods and compare them to the relevant state-of-the-art methods on both synthetic test functions and real-world benchmark problems.

\paragraph{Algorithms} 
We compare our algorithm to several baseline algorithms described in Section~\ref{sec:related_work} that can be applied to binary constraints: CEI \citep{gelbart2014bayesian}, SCBO \citep{2021Scalable}, SVM-CBO \citep{antonio2021sequential}, and random search. We implement and compare CEI and SCBO in our Python codebase, built upon the BoTorch \citep{balandat2020botorch} BO framework. We conduct SVM-CBO experiments using their framework. Our code will be released with a reproducibility guarantee. See more details in Appendix~\ref{app:setup:baseline}.

\paragraph{Benchmark Problems}
Our benchmark includes three synthetic test functions and nine real-world engineering design problems: 2D Townsend function \citep{townsend2014constrained}, 2D Simionescu function \citep{simionescu2014computer}, 2D LSQ function \citep{gramacy2016modeling}, 2D three-bar truss design \citep{ray2001engineering}, 3D tension-compression string design \citep{hedar2006derivative}, 4D welded beam design \citep{hedar2006derivative}, 4D gas transmission compressor design \citep{pant2009optimization}, 4D pressure vessel design \citep{coello2002constraint}, 7D speed reducer design \citep{lemonge2010constrained}, 9D planetary gear train design \citep{rao2012advanced}, 10D rolling element bearing design \citep{gupta2007multi}, and 30D cantilever beam design \citep{cheng2018adaptive}. 
% Note that we aim to solve the problem where the constraints are unknown and the algorithm only learns whether a design point is feasible or infeasible. Hence, multiple constraints can exist, but they are all captured with one classifier that outputs a binary value. 
Please refer to Appendix~\ref{app:setup:problems} for more detailed descriptions.

\begin{figure*}[t]
    \centering
    \begin{minipage}{0.92\textwidth}
     \begin{subfigure}[c]{0.28\textwidth}
        \centering
         \includegraphics[width=\textwidth]{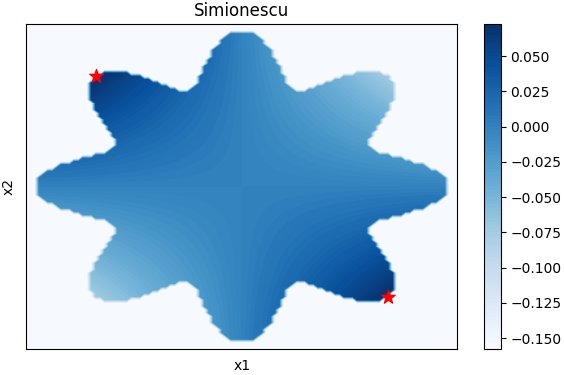}
     \end{subfigure}
     \hfill
     \begin{subfigure}[c]{0.7\textwidth}
        \centering
         \includegraphics[width=\textwidth]{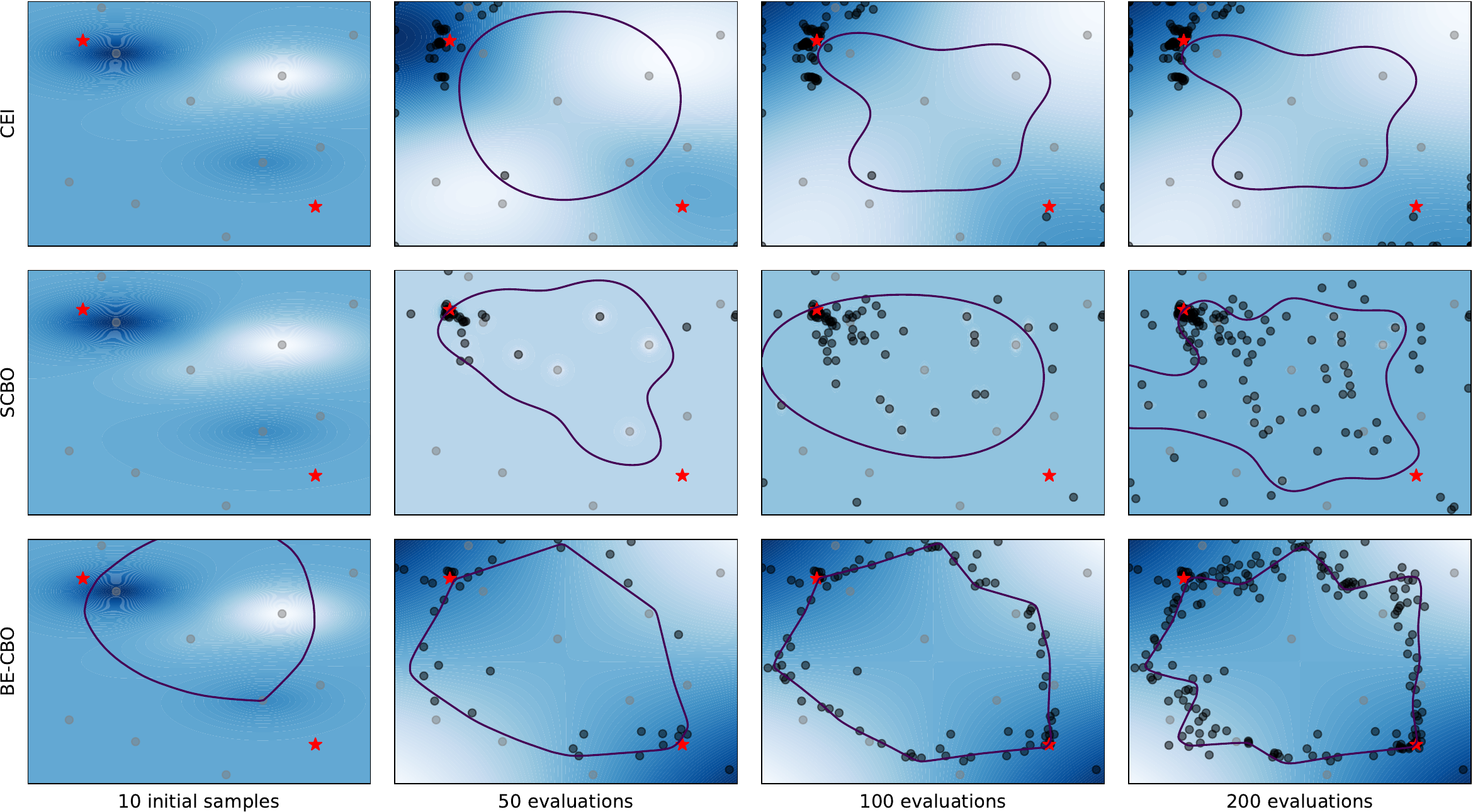}
     \end{subfigure}
     \end{minipage}
     \caption{Qualitative comparison of sample distributions from different algorithms on the Simionescu benchmark. Left: The true function landscape, where darker color means a higher objective value and the white region is infeasible. Right: The predicted function landscape (top: CEI, middle: SCBO\protect\footnotemark, bottom: BE-CBO) where darker color means a higher objective value and the contour represents the feasibility boundary (feasible inside, infeasible outside). Grey: initial samples, black: evaluated samples guided by each algorithm, and red: the global optima.}
     \label{fig:bdry_comp_algo}
\end{figure*}

\subsection{Results and Discussion}
\label{sec:discussion}

\paragraph{Performance} We monitor the current best value $f'$ over the number of evaluations performed. For every algorithm, we run experiments with 10 different random seeds and the same 10 initial samples for a total of 200 evaluations. We average the results over these 10 experiments for every algorithm, and plot the mean and the standard deviation across them, as presented in Figure~\ref{fig:performance}. 

Test problems are chosen to assess the efficacy of the method in handling a wide range of function characteristics, including concave, convex, disconnected, and varying complexities in design space. As shown in Figure~\ref{fig:performance}, BE-CBO consistently exhibits top performance and low variance compared to other algorithms that oscillates across different problems. Figure~\ref{fig:bdry_comp_algo} shows qualitative comparisons on sample distributions of algorithms at different time steps when evaluating on the Simionescu function. Our method effectively explores the boundary region, classifies the complex constraint landscape well, and discovers both optima of the function. Besides, our method also places samples occasionally on other parts of the boundary to capture other local optima that are potentially better than the best discovered optima. See more qualitative examples in Appendix~\ref{app:comp:qualitative}. We benchmark algorithm runtime in Appendix~\ref{app:comp:runtime}, which shows that BE-CBO exhibits stable runtime across all problem dimensions.

Overall, BE-CBO performs robustly on all of our benchmark problems, and we have not observed a case in which it completely fails or performs much worse than other baselines. However, we noticed that BE-CBO is slightly outperformed by SCBO on our 30D cantilever beam design problem. We speculate that this is because SCBO’s key advantage in high-dimensional optimization is its usage of TURBO~\cite{eriksson2019scalable}, a well-designed optimizer for high-dimensional BO based on the idea of using several local surrogates instead of a single global surrogate, along with its restart strategy when TURBO gets stuck. In BE-CBO, we adopt a relatively standard SLSQP optimizer with a global surrogate for acquisition optimization, which may be suboptimal for practical high-dimensional optimization. Although our core idea of BE-CBO is independent of the choice of acquisition optimizers, we believe that integrating effective ones such as TURBO with BE-CBO is technically feasible and promising for combining the best of both worlds.

\begin{figure*}
    \centering
    \includegraphics[width=\textwidth]{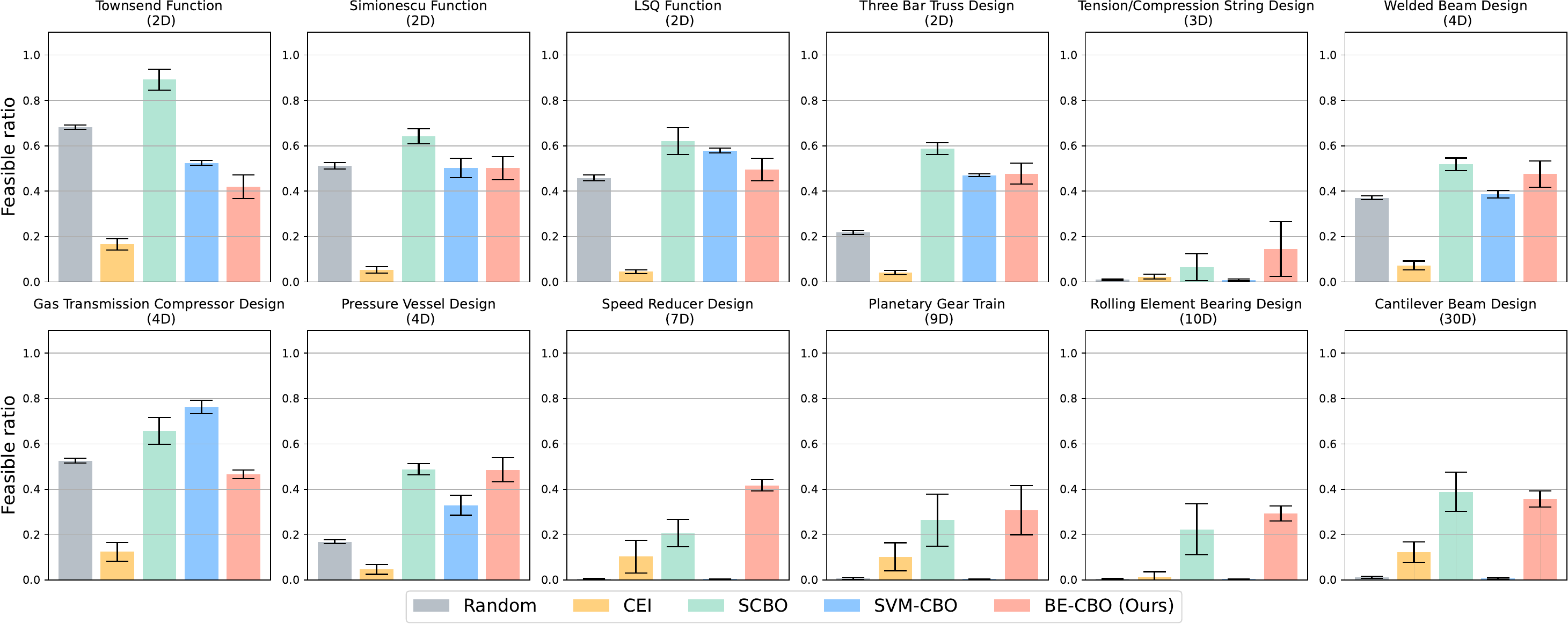}
    \caption{
        Feasibility ratio comparison of different algorithms including our BE-CBO on synthetic test functions and real-world problems. The error bar charts represent the mean and variance over 10 different random seeds.}
    \label{fig:feasibles}
\end{figure*}

\paragraph{Feasibility Ratio} We measure the \emph{feasibility ratio} in Figure~\ref{fig:feasibles} by comparing the number of feasible and infeasible points sampled by each algorithm. This is to keep track of how many samples did not retrieve any information on the objective value due to falling in the infeasible region. These samples do not improve the objective surrogate model, but they do improve the classifier and push the boundary further.

While the feasibility ratio has not been studied in CBO, there is abundant research on constrained evolutionary algorithms that focuses the search around the boundary and demonstrate improved performance~\citep{isaacs2008blessings, ray2009infeasibility, jiao2019feasible}. 
\citet{jiao2019feasible} empirically shows that 50\% is the most reliable feasibility ratio for evolutionary algorithms on global optimization problems with different function characteristics.

Interestingly, our algorithm outperforms the ones focusing the search in the feasible region (SCBO) or penalizing points less likely to be feasible (CEI).
%while still having a smaller ratio of feasibility in evaluations than SCBO. 
It is worth pointing to the examples where our algorithm is able to discover larger feasible regions and better performing designs due to the improved accuracy in constraint modeling and narrowed search around the boundary on the infeasible side. 

\footnotetext{Note that SCBO applies a Gaussian copula transform to the objective function to magnify the optimum region.}

\paragraph{Boundary Optimality} We focus on practical problems where the constraints are unknown and imposed by a physical system. The design keeps improving until it reaches the limits of the physical system. Hence, the optimum is often found on the boundary of feasible region. In the prior literature, this observation has been exploited in multiple contexts, such as designing effective constrained evolutionary algorithms~\citep{isaacs2008blessings, ray2009infeasibility, liu2021handling}, designing test functions that resemble real-world problems~\citep{sergeyev2021generator}, and active set methods in constrained optimization~\citep{gratton2011active}.

We acknowledge that this property does not hold for every problem, such as synthetically constructed functions with interior optima and problems with user-specified constraints.
Therefore, to test the generality of our method, we modified the three synthetic benchmark problems such that their optima are shifted to inside the feasible region. Results in Appendix~\ref{app:comp:interior} shows that BE-CBO successfully reaches the optimal region and exhibits competitive performance.

In addition, we investigate if active constraints exist at the global optima of all nine real-world problems in our benchmark, which cover most of the mechanical design problems with continuous parameters and unknown inequality constraints collected in a real-world non-convex constrained problem suite~\citep{kumar2020test}. For problems that do not have a known global optimum, we run CMA-ES~\citep{hansen2003reducing} with millions of evaluations to approximate a global optimum. As a result, we found that all nine problems have the global optimum on the constraint boundary, i.e., at least one active constraint. In those scenarios, the active constraints represent physical limitations that determine the upper bound of the mechanical design's performance. 
See more details of these experiments in Appendix~\ref{app:setup:problems:optimality}.

\subsection{Ablation Studies}

To systematically study how effective each component of BE-CBO is, we conduct comprehensive ablation study experiments around the two key contributions of BE-CBO: Deep Ensembles (DE) and boundary exploration. Please see Appendix~\ref{app:ablation} for details of each experiment. We aim to answer the following key questions:

\textit{1. How does DE improve classification accuracy upon GP?}
In Appendix~\ref{app:ablation:gp_de:cls_acc}, we compare the classification performance between DE and GP when being used as the classifier in BE-CBO. By measuring the \textit{Balanced Accuracy} metric from 10K random samples in the space in each iteration, we observe DE's superior and stable performance compared to GP on all benchmark problems.

\textit{2. How does DE improve BE-CBO’s performance upon GP?}
Appendix~\ref{app:ablation:gp_de:ours} shows that by switching DE to GP in BE-CBO, we observe significant performance drop on half of the benchmark problems. It confirms that having more accurate constraint classification helps improve the BO performance, especially for our boundary exploration strategy where boundary accuracy is important, and for real physical problems where boundary is crucial to explore.

\textit{3. Is DE sensitive to hyper-parameters?}
We find that the overall performance of DE is insensitive to hyperparameters within wide ranges by experimenting different numbers of MLPs, numbers of hidden layers, numbers of neurons in a layer, and learning rates in Appendix~\ref{app:ablation:de}.

\textit{4. How does BE-CBO perform with other acquisition functions than EI?} We use EI as the acquisition function in BE-CBO mainly because of its popularity and good balance between exploration and exploitation. However, BE-CBO is compatible with any standard acquisition functions and is not tied to EI. One can also use upper confidence bound (UCB) when more explicit control over the exploration is preferred. In Appendix~\ref{app:ablation:acq}, we compare the performance of using EI and UCB for BE-CBO respectively, and the results suggest that they perform similarly well on our benchmark problems overall with some differences in particular problems.

\textit{5. How does boundary exploration perform compared to other forms of constrained acquisition functions?}
We investigate two variants of BE-CBO where similar constraint treatments to CEI and SCBO are applied instead of the proposed boundary exploration. In Appendix~\ref{app:ablation:be}, the results show that our boundary exploration is the most robust strategy in coupling the constraints with the acquisition function.
\section{Conclusion and Future Work}
We introduced a novel CBO method that aims to find a global optimum of an unknown function under unknown constraints. We specifically address the problem in which the infeasible designs cannot be evaluated, hence returning no information for the objective function nor a continuous value for the constraint. 
Such problems are common, for example, in chemistry and materials science in which specific combinations of synthesis parameters may lead to invalid materials, making it impossible to measure any output quantities. This results in a BO iteration with no information gain which can cause the optimizer to get stuck in the worst case. 
Coupling the BO optimizer with a classifier that can distinguish the feasible from the infeasible regions allows the optimizer to also draw information from failed experiments and to converge quickly to the actual optimum most often located at the constrained boundary, paving the way to more efficient experimentation. 

We employed \textit{Deep Ensembles} for the classification of the constraints representing binary feasible/infeasible regions. To the best of our knowledge, we are the first to propose using DE for modeling unknown constraints in BO. DE exhibits improved accuracy in the classification, allowing our method to focus the search for optima on the boundary. However, we share the same observation as \citet{li2023study} that using DE for modeling objectives (i.e., regression) does not lead to improved performance. Finally, we present a boundary exploration strategy that efficiently discovers better designs. We performed extensive tests on both synthetic test functions and real-world problems. We found that our approach outperforms other methods and works particularly well on practical problems.

We acknowledge two main limitations of our work. Firstly, our algorithm was exclusively tested in a continuous design space, focusing on a single objective. To broaden its applicability, a future direction would involve extending the algorithm to handle categorical variables and explore multi-objective BO. Secondly, our work does not account for the potential costs linked to evaluating infeasible samples. For instance, costs associated with material breakage during evaluation or motor damage caused by excessively high voltage are not considered. Furthermore, to the best of our knowledge, we are unable to find theoretical analysis on global convergence rate in the constrained BO literature when unknown constraints are involved due to the added complexity of constraint surrogates and the interplay between constraints and objectives.
We leave these exciting directions for future work.

\section*{Acknowledgements}
We thank the reviewers for their constructive suggestions. This work is supported by the MIT-IBM Watson AI Lab, and its member company, Evonik. This work also received the support of a fellowship from “la Caixa” Foundation (ID 100010434). The fellowship code is LCF/BQ/EU21/11890103.

\section*{Impact Statement}
This paper presents work whose goal is to advance the field of Machine Learning. There are many potential societal consequences of our work, none which we feel must be specifically highlighted here.

\bibliography{references}
\bibliographystyle{icml2024}

%%%%%%%%%%%%%%%%%%%%%%%%%%%%%%%%%%%%%%%%%%%%%%%%%%%%%%%%%%%%%%%%%%%%%%%%%%%%%%%
%%%%%%%%%%%%%%%%%%%%%%%%%%%%%%%%%%%%%%%%%%%%%%%%%%%%%%%%%%%%%%%%%%%%%%%%%%%%%%%
% APPENDIX
%%%%%%%%%%%%%%%%%%%%%%%%%%%%%%%%%%%%%%%%%%%%%%%%%%%%%%%%%%%%%%%%%%%%%%%%%%%%%%%
%%%%%%%%%%%%%%%%%%%%%%%%%%%%%%%%%%%%%%%%%%%%%%%%%%%%%%%%%%%%%%%%%%%%%%%%%%%%%%%
\newpage
\appendix
\onecolumn

\section{Method Implementation Details}
\label{app:method}

\subsection{Deep Ensembles}
\label{app:method:de}

\subsubsection{Neural Network Architecture and Parameters}
\label{app:method:de:nn_arch}

We implement an ensemble of Multi-layer Perceptrons (MLPs) for modeling the unknown constraints. For each MLP in the ensemble, we use a simple and standard structure of 4 fully connected layers with $64\lfloor\log_2(d)\rfloor$ neurons in each hidden layer where $d$ is the problem dimension. The network gets larger as the problem dimension gets higher. We use ReLU nonlinearity between each pair of fully connected layers.

Similar to training GP classifiers using a variational framework~\citep{hensman2015scalable}, we use the variational evidence lower bound (ELBO) to approximate the posterior and marginal likelihood of \textit{Deep Ensemble} classifiers given the Bernoulli likelihood. The ensemble is optimized for maximal marginal log likelihood using the Adam optimizer with a $3\times10^{-4}$ learning rate for 1,000 iterations. Note that we do not apply regularization or dropout as suggested by \citet{lakshminarayanan2017simple}. 
% The ensemble is trained until reaching 100\% accuracy on the training data or 1,000 iterations, whichever comes earlier.

Overall, the architecture we use is very simple and straightforward to implement with almost no hyper-parameters, but works robustly across a wide range of benchmark problems as shown in our experiments.

\subsubsection{Mean and Uncertainty Computation}
\label{app:method:de:mean_std}

In Section 4.1 of the main paper, we introduced the mean and variance computation following the original formulation of \textit{Deep Ensembles}~\citep{lakshminarayanan2017simple}. In practice, we empirically observed an alternative implementation of mean and uncertainty computation leads to better performance. 

As mentioned in Section~\ref{classifier}, instead of letting the network ensemble directly output the probability value between $[0, 1]$ for the constraint feasibility and training the ensemble using maximum likelihood estimation (MLE, specifically, binary cross-entropy loss), we treat the network ensemble as a \textit{real-valued latent constraint function} $g(x)$ such that the constraint is satisfied if and only if $g(x)\geq 0$. In other words, we let the network ensemble output a latent Gaussian distribution in a continuous space and later transform the output to a probability between $[0, 1]$ by standard normal CDF $\Phi$, following the approach taken by CEI~\citep{gelbart2014bayesian}. The transformed probability is essentially the mean prediction. With such treatment, we can optimize the \textit{Deep Ensembles} in a similar way as training GP classifiers based on a variational inference (VI) framework, as described in Section~\ref{app:method:de:nn_arch}.

For uncertainty computation, since we can easily get the real-valued mean $\mu_g$ and standard deviation $\sigma_g$ from the latent Gaussian distribution, we can transform the one-sigma confidence interval $[\mu_g - \sigma_g, \mu_g + \sigma_g]$ by the standard normal CDF $\Phi$ and obtain a corresponding confidence interval in the transformed probability space $[\Phi(\mu_g - \sigma_g), \Phi(\mu_g + \sigma_g)]$ and define the transformed standard deviation as $\sigma_E = (\Phi(\mu_g + \sigma_g) - \Phi(\mu_g - \sigma_g)) / 2$. In practice, we use this formula to obtain the dynamic bounds as described in Section~\ref{sec:be} of the main paper.

\begin{figure}[h]
    \centering
    \includegraphics[width=1.0\textwidth]{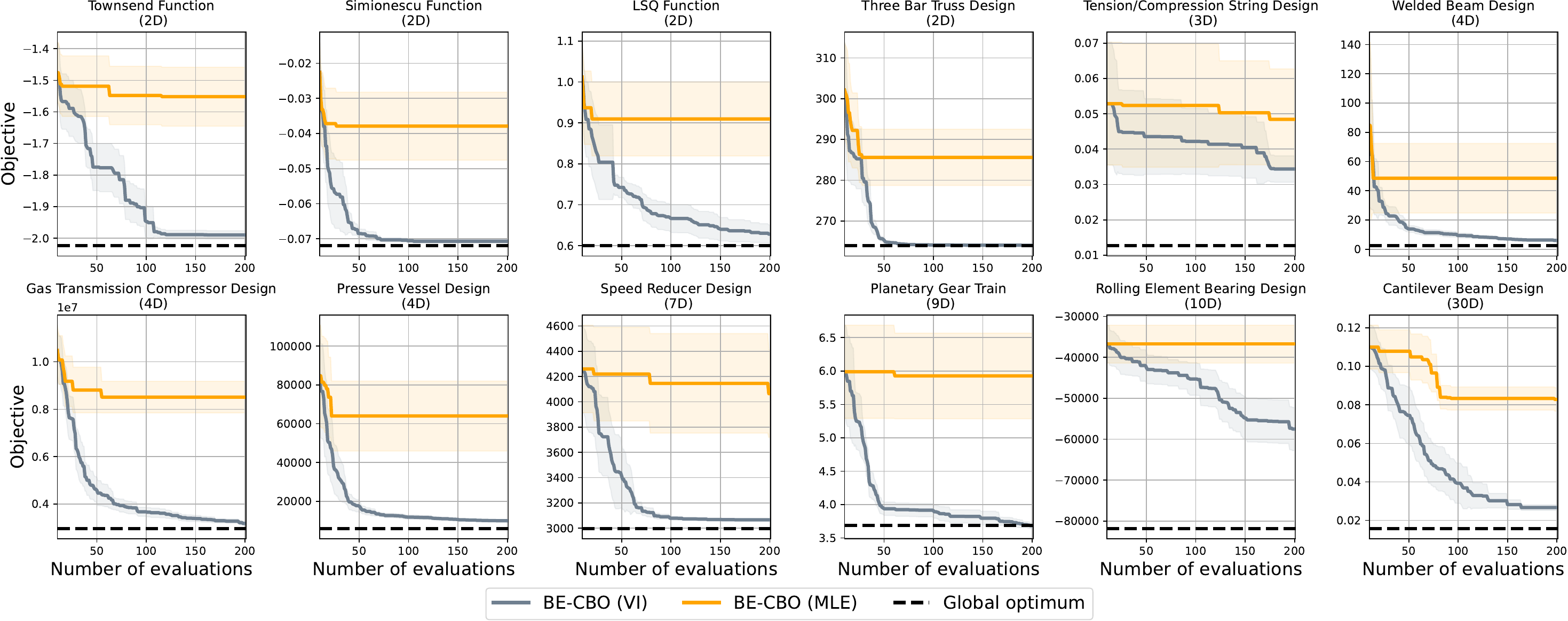}
    \caption{Comparisons between BE-CBO trained with VI and MLE respectively, averaged over 10 random seeds.}
    \label{fig:vi_mle}
\end{figure}

In practice, we find that models trained by MLE are overly confident, i.e., usually output extremely low uncertainties, which leads to the poor performance. On the contrary, training with VI produces much more reasonable uncertainty estimation. Figure~\ref{fig:vi_mle} shows empirical comparisons between training DE with VI and MLE respectively for BE-CBO, which shows a clear advantage of VI.

\subsection{Constrained Optimization on Acquisition Functions}

For numerical optimization of the constrained acquisition function in BE-CBO, we specify the dynamic bound constraint as a nonlinear inequality constraint to the acquisition optimizer in BoTorch~\citep{balandat2020botorch}, which calls an underlying SLSQP optimization from SciPy, a standard and robust choice for handling nonlinear inequality constraints. However, the SLSQP optimizer requires constraint-satisfying initial solutions as input. Since randomly generated initial solutions can easily violate the constraint, to deal with this problem, we use the Adam optimizer \citep{kingma2014adam} to find constraint-satisfying samples through gradient descent. We use a standard mean squared error (MSE) loss to measure the difference between the sample's feasibility probability and $0.5$. By optimizing such MSE loss, the sample gets closer to the constraint boundary by pushing the feasibility probability to $0.5$, thus the sample becomes closer to satisfying the dynamic band constraint.

\section{Experimental Setup}
\label{app:setup}

Due to the large number of benchmark problems and random seeds, the experiments are conducted in parallel on a distributed server with Intel Xeon Platinum 8260 CPUs with 4GB RAM per core, where each individual experiment runs on a single CPU thread without GPU.

For surrogate modeling of the objective in BE-CBO and all baseline algorithms, we directly use the implementation of GP regressors in BoTorch~\citep{balandat2020botorch}, where Matern 5/2 kernel is used with their default hyperparameters\footnote[1]{\url{https://botorch.org/tutorials/scalable_constrained_bo}}. For the GP constraint classifiers in all baseline algorithms and also ablation studies in BE-CBO, we leverage the implementation from GPyTorch~\citep{gardner2018gpytorch}, where RBF kernel is used with their default hyperparameters.\footnote[2]{ \url{https://notebook.community/jrg365/gpytorch/examples/02_Simple_GP_Classification/Simple_GP_Classification}} We use the standard Bernoulli likelihood in GP classifiers for representing posterior probability distributions, and use the variational evidence lower bound (ELBO) to optimize the GP classifiers~\citep{hensman2015scalable}.

\subsection{Benchmark Problems}
\label{app:setup:problems}
In this section, we briefly introduce the properties of each problem, including the dimensions of the design space $\mathcal{X} \subset \mathbb{R}^d $. The problem descriptions for 3 synthetic functions and 9 real-world problems are described respectively. We perform 10 independent test runs with 10 different random seeds for each problem on each algorithm. For each test run of one problem, we use the same initial set of samples for every algorithm, which is generated by a scrambled quasirandom Sobol sequence using the same random seed.

We represents all benchmark problems in the form of minimizing $f(x)$ subject to multiple underlying constraints $c_1(x)\geq 0,...,c_n(x)\geq 0$. Note that we aim to solve the problem where the constraints are unknown and the algorithm only learns whether a design point is feasible or infeasible. Hence, multiple constraints can exist, but their formulas are invisible and they are all captured with one classifier that outputs a binary value. 

\subsubsection{Synthetic Test Functions}
\paragraph{Townsend function~\citep{townsend2014constrained}} is a trigonometric function $f(x) = - [\cos((x_1 - 0.1)x_2)]^2 - x_1 \sin(3x_1 + x_2)$ constrained by $c(x) = (2\cos t - \frac{1}{2}\cos 2t - \frac{1}{4}\cos 3t - \frac{1}{8}\cos 4t )^2 + (2\sin t)^2 - x_1^2 - x_2^2$ where $t = \arctantwo (x_1, x_2)$ and with bounds $-2.25 \le x_1 \le 2.25$ and $-2.5 \le x_2 \le 1.75$.
    
\paragraph{Simionescu function~\citep{simionescu2014computer}} is a hyperbolic paraboloid function $f(x) = 0.1x_1x_2$ constrained by $c(x) = (r_T + r_S\cos(n \arctan \frac{x}{y}))^2 - x_1^2 - x_2^2$ where $r_T = 1$, $r_S = 0.2$ and $n = 8$, in the domain $[-1.25, 1.25]^2$.
    
\paragraph{LSQ function~\citep{gramacy2016modeling}} is a linear objective function $f(x) = x_1 + x_2$ with sinusoidal and quadratic constraints $c_1(x) = x_1 + 2x_2 + \frac{1}{2}\sin(2\pi(x_1^2-2x_2)) - \frac{3}{2}$ and $c_2(x) = \frac{3}{2} - x_1^2 - x_2^2$ bounded by $[0, 1]^2$.

\subsubsection{Real Test Functions}

\paragraph{Three bar truss design~\citep{ray2001engineering}} minimizes the volume of the truss structure subject to stress constraints. The analytical formula is given as $f(x) = l(2\sqrt{2}x_1 + x_2)$ with two variables $0 \leq x_1, x_2 \leq 1$, subject to three constraints:
\begin{multicols}{3}
\begin{itemize}
    \item $c_1(x) = 2 - \frac{\sqrt{2x_1} + x_2}{\sqrt{2}x_1^2 + 2x_1x_2}$
    \item $c_2(x) = 2 - \frac{1}{x_1 + \sqrt{2}x_2}$
    \item $c_3(x) = 2 - \frac{x_2}{\sqrt{2}x_1^2 + 2x_1x_2}$
\end{itemize}
\end{multicols}

\paragraph{Tension-compression string design~\citep{hedar2006derivative}} is a three dimensional design problem where the weight of a tension-compression string, given by $f(x) = (x_1 + 2)x_2x_3^2$, needs to be minimized, where: $2 \le x_1 \le 15$ is the number of active coils (integer), $0.25 \le x_2 \le 1.3$ is the wire diameter, and $0.05 \le x_3 \le 2$ is the mean coil diameter. This minimization is constrained by the minimum deflection, shear stress, surge frequency, and limits on the outside diameter \citep{coello2002constraint}:
\begin{multicols}{2}
\begin{itemize}
    \item $c_1(x) = \frac{x_2^3x_1}{71785x_3^4} - 1$
    \item $c_2(x) = 1 - \frac{4x_2^2 - x_3x_2}{12566(x_2x_3^3 - x_3^4)} - \frac{1}{5 108x_3^2}$
    \item $c_3(x) = \frac{140.45x_3}{x_2^2x_1} - 1$
    \item $c_4(x) = 1 - \frac{x_2 + x_3}{1.5}$
\end{itemize}
\end{multicols}

\paragraph{Welded beam design~\citep{belegundu1985study}} aims to minimize the cost of the beam, $f(x) = 1.10471x_1^2x_2 + 0.04811x_3x_4(14 + x_2)$, where $0.125 \le x_1 \le 10$ and $0.1 \le x_2, x_3, x_4 \le 10$ are four design variables referring to physical dimensions of the beam. This optimization is subject to constraints on the shear stress, bending stress in the beam, buckling load on the bar, the end deflection of the beam, and a side constraint \citep{hedar2006derivative}:
\begin{multicols}{3}
\begin{itemize}
    \item $c_1(x) = 13000 - \tau(x) $
    \item $c_2(x) = 30000 - \sigma(x)$
    \item $c_3(x) = P_c(x) - 6000$
    \item $c_4(x) = 0.25 - \delta(x)$
    \item $c_5(x) = x_4 - x_1$
\end{itemize}
\end{multicols}
where $\tau(x) = \sqrt{(\tau_1(x))^2 + (\tau_2(x))^2 + \frac{x_2\tau_1(x)\tau_2(x)}{\sqrt{0.25[x_2^2 + (x_1 + x_3)^2]}}}$, $\tau_1(x) = \frac{6000}{\sqrt{2}x_1x_2}$, $\tau_2(x) = \frac{6000(14 + 0.5x_2) \sqrt{0.25[x_2^2 + (x_1 + x_3)^2]}}{2[0.707x_1x_2(x_2^2/12 + 0.25(x_1 + x_3)^2)]}$, $\sigma(x) = \frac{504000}{x_3^2x_4}$, $P_c(x) = 64746.022(1 - 0.0282346x_3)x_3x_4^3$, $\delta(x) = \frac{2.1953}{x_3^3x_4}$.

\paragraph{Gas transmission compressor design~\citep{pant2009optimization}} aims to minimize the total annual cost of a gas pipeline transmission system and its operation, given by $f(x) = (8.61)10^5x_1^{1/2}x_2x_3^{-2/3}x_4^{-1/2} + (3.69)10^4x_3 + (7.72)10^8x_1^{-1}x_2^{0.219} - (765.43)10^6x_1^{-1}$, where $20 \le x_1 \le 50$ is the length between compressor stations, $1 \le x_2 \le 10$ is the compression ratio, $ 20 \le x_3 \le 50$ is the inside diameter of the pipe, and $0.1 \le x_4 \le 60$ is a non-dimensional parameter, subject to $c(x) = 1 - x_4x_2^{-2} - x_2^{-1}$.

\paragraph{Pressure vessel design~\citep{coello2002constraint}} minimize the total cost $f(x) = 0.6224x_1x_3x_4 + 1.7781x_2x_3^2 + 3.1661x_1^2x_4 + 19.84x_1^2x_3$, including the cost of the material, forming and welding. There are four design variables: Ts (thickness of the shell), Th (thickness of the head), R (inner radius) and L (length of the cylindrical section of the vessel, not including the head). The constraints are given as:
\begin{multicols}{2}
\begin{itemize}
    \item $c_1(x) = x1 - 0.0193x_3$
    \item $c_2(x) = x2 - 0.00954x_3$
    \item $c_3(x) = \pi x_3^2x_4 + \frac{4}{3}\pi x_3^3 - 1296000$
    \item $c_4(x) = 240 - x_4$
\end{itemize}
\end{multicols}

\paragraph{Speed reducer design~\citep{lemonge2010constrained}} is a seven dimensional problem that seeks to minimize the weight of a speed reducer. The design variables are: the face width ($2.6 \le x_1 \le 3.6$), the module of teeth ($0.7 \le x_2 \le 0.8$), the number of teeth on pinion (integer) ($17 \le x_3 \le 28$), the length of the shaft 1 between the bearings ($7.3 \le x_4 \le 8.3$), the length of the shaft 2 between the bearings ($7.3 \le x_5 \le 8.3$), the diameter of the shaft 1 ($2.9 \le x_6 \le 3.9$), and the diameter of the shaft 2 ($5 \le x_7 \le 5.5$).\par
The weight is given by:\par
$f(x) = 0.7854x_1x_2^2(3.3333x_3^2 + 14.9334x_3 - 43.0934) - 1.508x_1(x_6^2 + x7^2) + 7.4777(x_6^3 + x_7^3) + 0.7854(x_4x_6^2 + x_5x_7^2)$\par
and is subject to the following mechanical constraints:
\begin{multicols}{2}
\begin{itemize}
    \item $c_1(x) = 1 - 27x_1^{-1}x_2^{-2}x_3^{-1}$,
    \item $c_2(x) = 1 - 397.5x_1^{-1}x_2^{-2}x_3^{-2}$,
    \item $c_3(x) = 1 - 1.93x_2^{-1}x_3^{-1}x_4^{-3}x_6^{-4}$,
    \item $c_4(x) = 1 - 1.93x_2^{-1}x_3^{-1}x_5^{-3}x_7^{-4}$,
    \item $c_5(x) = 1100 - [\frac{745x_4}{x_2x_3}^2 + (16.9)10^6]^{0.5}\frac{1}{0.1x_6^3}$,
    \item $c_6(x) = 850 - [\frac{745x_5}{x_2x_3}^2 + (157.5)10^6]^{0.5}\frac{1}{0.1x_7^3}$,
    \item $c_7(x) = 40 - x_2x_3$,
    \item $c_8(x) = x_1/x_2 - 5$,
    \item $c_9(x) = 12 - x_1/x_2$,
    \item $c_{10}(x) = 1 - (1.5x_6 + 1.9)x_4^{-1}$,
    \item $c_{11}(x) = 1 - (1.1x_7 + 1.9)x_5^{-1}$.
\end{itemize}
\end{multicols}

\paragraph{Planetary gear train design~\citep{rao2012advanced}} minimizes the gear ratio errors which can be stated as:\par 
$f(x) = \max | i_k - i_{0k} |$ where $k = \{1, 2, R\}$, $i_1 = \frac{N_6}{N_4}$, $i_{01} = 3.11$, $i_2 = \frac{N_6(N_1N_3 + N_2N_4)}{N_1N_3(N_6 - N_4)}$, $i_{02} = 1.84$, $i_R = \left( \frac{N_2N_6}{N_1N_3} \right)$, $i_{0R} = -3.11$. The design variables are defined as $\bar{x} = (\rho, N_6, N_5, N_4, N_3, N_2, N_1, m_2, m_1)$. It is also subject to the following constraints:
\begin{multicols}{2}
\begin{itemize}
    \item \( c_1(x) = D_{\max} - m_3(N_6 + 2.5) \)
    \item \( c_2(x) = D_{\max} - m_1(N_1 + N_2) - m_1(N_2 + 2) \)
    \item \( c_3(x) = D_{\max} - m_3(N_4 + N_5) - m_3(N_5 + 2) \)
    \item \( c_4(x) = (N_1 + N_2)\sin\left(\frac{7}{\rho}\right) - N_2 - 2 - \delta_{22} \)
    \item \( c_5(x) = (N_6 - N_3)\sin\left(\frac{7}{\rho}\right) - N_3 - 2 - \delta_{33} \)
    \item \( c_6(x) = (N_4 + N_5)\sin\left(\frac{7}{\rho}\right) - N_5 - 2 - \delta_{55} \)
    \item \( c_7(x) = (N_6 - N_3)^2 - (N_3 + N_5 + 2 + \delta_{35})^2 - (N_4 + N_5)^2 - 2(N_6 - N_3)(N_4 + N_5)\cos\left(\frac{2\pi}{\rho} - \beta\right) \)
    \item \( c_8(x) = -N_4 + N_6 - 2N_5 - 2\delta_{56} - 4 \)
    \item \( c_9(x) = -2N_3 + N_6 - N_4 + 2\delta_{34} + 4 \)
\end{itemize}
\end{multicols}
where \( \delta_{22} = \delta_{33} = \delta_{55} = \delta_{35} = \delta_{56} = 0.5 \), \( \beta = \cos^{-1}\left(\frac{(N_4 + N_5)^2 + (N_6 - N_3)^2 - (N_3 + N_5)^2}{2(N_6 - N_3)(N_4 + N_5)}\right) \), and \( D_{\max} = 220 \), with bounds: \( p = (3, 4, 5) \), \( m_1 = (1.75, 2.0, 2.25, 2.5, 2.75, 3.0) \), \( m_3 = (1.75, 2.0, 2.25, 2.5, 2.75, 3.0) \), \( 17 \leq N_1 \leq 96 \), \( 14 \leq N_2 \leq 54 \), \( 14 \leq N_3 \leq 51 \), \( 17 \leq N_4 \leq 46 \), \( 14 \leq N_5 \leq 51 \), \( 48 \leq N_6 \leq 124 \), and \( N_i \) is an integer. We apply continuous optimization methods on this problem by rounding the continuous variables into integer ones.

\paragraph{Rolling element bearing design~\citep{gupta2007multi}} optimizes the dynamic capacity of a rolling bearing. The design parameter vector can be written as $\bar{x} = (D_m, D_b, Z, f_i, f_o, K_{\text{Dmin}}, K_{\text{Dmax}}, \varepsilon, e, \xi)$ where $\bar{f}_i = \frac{r_i}{D_b}$ and $\bar{f}_o = \frac{r_o}{D_b}$. The bounds of the paramters are \( D_m \in [125, 150] \), \( D_b \in [10.5, 31.5] \), \( Z \in [4, 50] \), \( f_i \in [0.515, 0.6] \), \( f_o \in [0.515, 0.6] \), \( K_{\text{Dmin}} \in [0.4, 0.5] \), \( K_{\text{Dmax}} \in [0.6, 0.7] \), \( \varepsilon \in [0.3, 0.4] \), \( e \in [0.02, 0.1] \), \( \xi \in [0.6, 0.85] \).
The objective function is defined as $f(x) = C_d$ where
\[
C_d = 
\begin{cases} 
f_cZ^{2/3}D_b^{1.8}, & D_b \leq 25.4 \text{ mm} \\
3.647f_cZ^{2/3}D_b^{1.4}, & D_b > 25.4 \text{ mm}
\end{cases}
\]
\[
f_c = 37.91 \left[ 1 + \left\{ 1.04 \left( \frac{(1-\gamma)^{1.72}}{(1+\gamma)^{1.72}} \right) \left( \frac{f_i(2f_o - 1)}{f_o(2f_i - 1)} \right)^{0.41} \right\}^{10/3^{0.3}} \right] \left[ \frac{\gamma^{0.3}(1 - \gamma)^{1.39}}{(1 + \gamma)^{1/3}} \right] \left[ \frac{2f_i}{2f_i - 1} \right]^{0.41},
\gamma = \frac{D_b \cos \alpha}{D_m}.
\]
The constraints are defined as:
\begin{multicols}{2}
\begin{itemize}
    \item \( c_1(x) = \frac{\phi_0}{2 \sin^{-1}\left(\frac{D_b}{D_m}\right)} - Z + 1 \)
    \item \( c_2(x) = 2D_b - K_{\text{Dmin}}(D - d) \)
    \item \( c_3(x) = K_{\text{Dmax}}(D - d) - 2D_b \)
    \item \( c_4(x) = \xi_{B_w} - D_b \)
    \item \( c_5(x) = D_m - 0.5(D + d) \)
    \item \( c_6(x) = (0.5 + e)(D + d) - D_m \)
    \item \( c_7(x) = 0.5(D - D_m - D_b) - eD_b \)
\end{itemize}
\end{multicols}
where 
\[
\phi_0 = 2\pi - 2\cos^{-1} \left[ \frac{\left\{ (D - d)/2 - 3(T/4)^2 \right\}^2 + \left\{ (D/2 - (T/4) - D_b) \right\}^2 - (d/2 + (T/4))^2}{2\{(D - d)/2 - 3(T/4)\}\{(D/2 - (T/4) - D_b)\}} \right], T = D - d - 2D_b.
\]

\paragraph{Cantilever beam design~\citep{cheng2018adaptive}} minimizes the tip deflection of a stepped cantilever beam subject to constraints. Due to the complexity of formulas, please refer to the original paper for details.

\subsubsection{Boundary Optimality of Real-World Problems}
\label{app:setup:problems:optimality}

\begin{table}[h!]
\caption{Details of the 9 real-world design optimization problems. $d$ is the dimension of design variables, $g$ is the number of unknown inequality constraints, $g^*$ is the number of active constraints at the optimum, $f(x^*)$ is the optimal performance value.}
\begin{center}
\begin{tabular}{lllll}
\toprule
    Problem Name & $d$ & $g$ & $g^*$ & $f(x^*)$ \\
\midrule
    Three-bar truss design problem & 2 & 3 & 1 & 2.6389E+02 \\
    Tension/compression spring design & 3 & 3 & 2 & 1.2665E-02 \\
    Welded beam design & 4 & 5 & 1 & 2.4453E+00 \\
    Gas transmission compressor design & 4 & 1 & 1 & 2.9648E+06 \\
    Pressure vessel design & 4 & 4 & 2 & 5.8853E+03 \\
    Speed reducer design & 7 & 11 & 2 & 2.9944E+03 \\
    Planetary gear train design & 9 & 9 & 1 & 3.6859E+00 \\
    Rolling element bearing & 10 & 9 & 3 & 8.3918E+04 \\
    Cantilever beam design & 30 & 21 & 1 & 1.5731E+02 \\
\bottomrule
\end{tabular}
\label{tab:problem_optimality}
\end{center}
\end{table}

For real-world design optimization problems with unknown physical constraints, the optimal solutions usually lie on the boundary of feasible regions since the performance upper bound is usually constrained by physical limits. Though this is often the experience from practitioners, to further validate whether this observation is true, we check whether the global optimum is on the constraint boundary for each problem in our collected nine benchmark problems. Table~\ref{tab:problem_optimality} summarizes the problem description, the optimum information and active constraints at the optimum. All problems have at least one active constraint at the optimum and some problems have multiple ones.

\subsection{Baseline Algorithms}
\label{app:setup:baseline}

In this section, we provide a description of the other baseline algorithms that we used for comparison with our own algorithm. Specifically, we compare our algorithms to several BO baseline algorithms that are discussed in Results Section of the main paper. These baseline algorithms are designed or have been modified to handle binary unknown constraints. Here we describe the differences between our adoption of these algorithms and their original formulation in the paper.

\paragraph{CEI~\citep{gelbart2014bayesian}} We implement this algorithm in BoTorch. The acquisition function remains unchanged to its original form, with one notable modification regarding the constraint classification. In contrast to the paper's setup where individual constraints provide separate responses, our evaluation process only allows for obtaining a single feasibility response. Therefore, we have modified the implementation to utilize a single binary classifier as a constraint surrogate instead of multiple classifiers for multiple constraints as described in the original paper. 

\paragraph{SCBO~\citep{2021Scalable}} We implement this algorithm in BoTorch as there exists a reference implementation in the BoTorch documentation\footnote[1]{\url{https://botorch.org/tutorials/scalable_constrained_bo}}. In contrast to the original setup, where multiple \textit{continuous} responses can be obtained from individual constraints during evaluation, our approach assumes a \textit{binary} feasibility response. Consequently, we employ a single binary classifier in our setup, similar to the adoption of CEI, instead of multiple regressors for multiple constraints as described in the paper. Due to the binary nature of the constraint, we do not apply the Bilog transformation to the constraint values. However, we still apply the Gaussian copula transformation to the objective values and implement the restart of new trust regions according to the paper's descriptions.

\paragraph{SVM-CBO~\citep{antonio2021sequential}}
To conduct the SVM-CBO algorithm experiments, we utilized the code and framework provided by the authors. The experimental setup described in their paper involved 100 evaluations, comprising of 10 initial samples, 60 evaluations in phase 1, and 10 evaluations in phase 2. In our extended evaluations, where we employed 20 initial samples and a total of 200 evaluations, we maintained the same ratio as described in the original paper. Thus, we performed 127 evaluations in phase 1 and 63 evaluations in phase 2.

\section{Additional Comparison}
\label{app:comp}

\subsection{Qualitative Comparison of Sample Distributions}
\label{app:comp:qualitative}

\begin{figure*}[h]
    \centering
    \begin{minipage}{\textwidth}
     \begin{subfigure}[c]{0.28\textwidth}
        \centering
         \includegraphics[width=\textwidth]{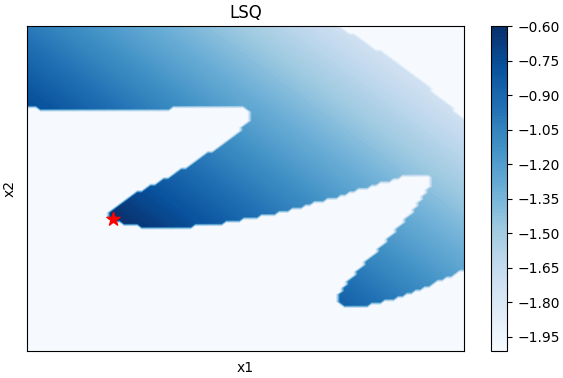}
     \end{subfigure}
     \hfill
     \begin{subfigure}[c]{0.7\textwidth}
        \centering
         \includegraphics[width=\textwidth]{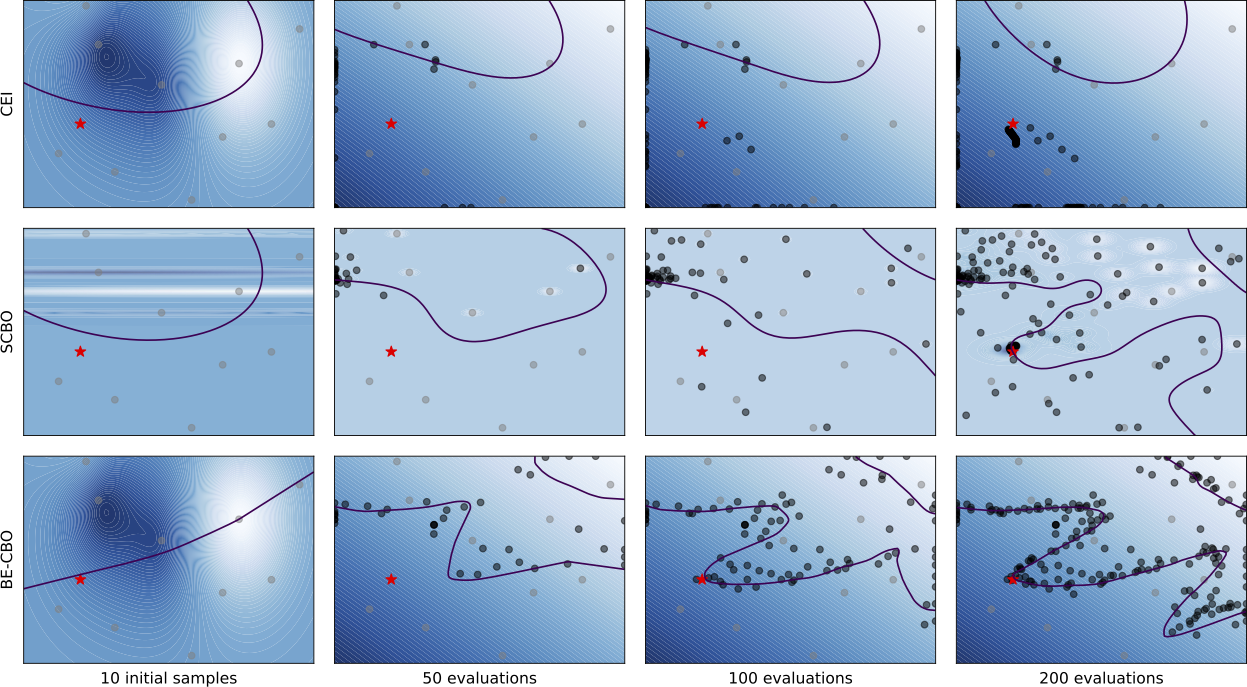}
     \end{subfigure}
     \end{minipage}
     \caption{Qualitative comparison of sample distributions from different algorithms on the LSQ benchmark. }
     \label{fig:bdry_comp_lsq}
\end{figure*}

\begin{figure*}[h]
    \centering
    \begin{minipage}{\textwidth}
     \begin{subfigure}[c]{0.28\textwidth}
        \centering
         \includegraphics[width=\textwidth]{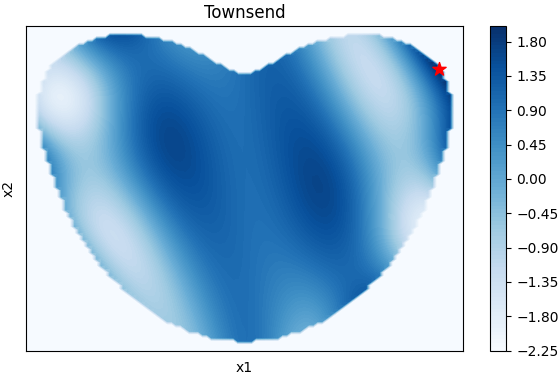}
     \end{subfigure}
     \hfill
     \begin{subfigure}[c]{0.7\textwidth}
        \centering
         \includegraphics[width=\textwidth]{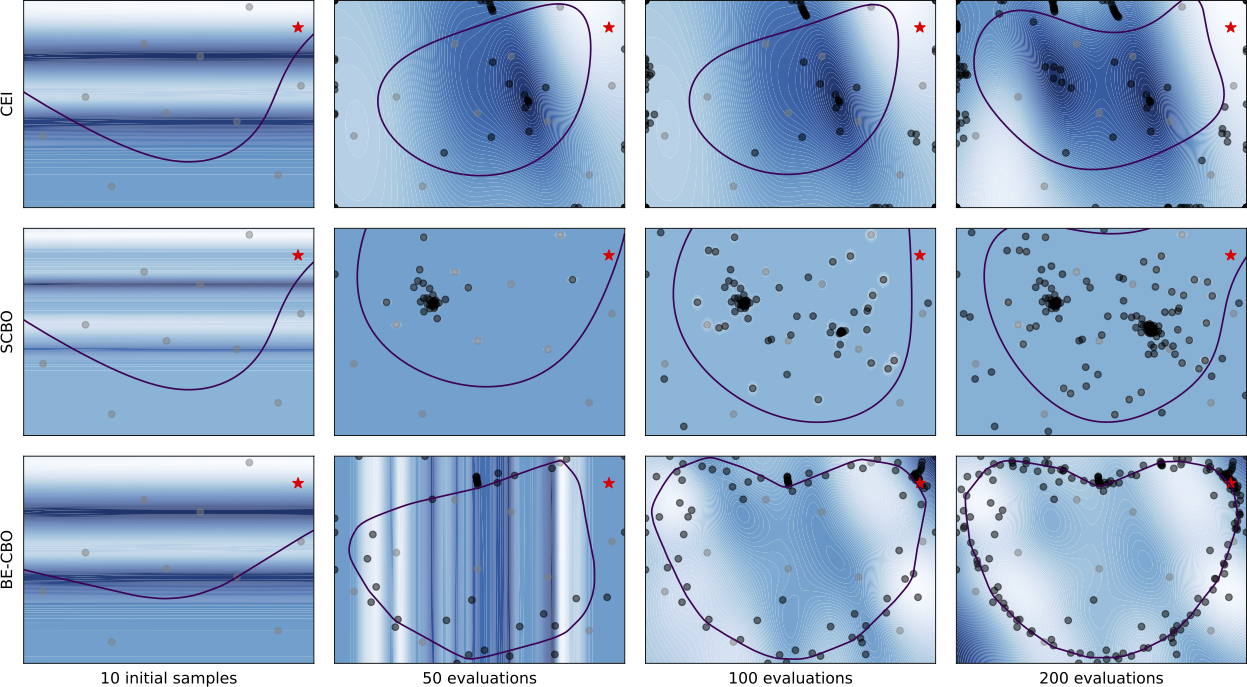}
     \end{subfigure}
     \end{minipage}
     \caption{Qualitative comparison of sample distributions from different algorithms on the Townsend benchmark. }
     \label{fig:bdry_comp_tow}
\end{figure*}

We show more qualitative comparison of sample distributions on extra two 2D functions (LSQ and Townsend) in Figure~\ref{fig:bdry_comp_lsq} and \ref{fig:bdry_comp_tow} besides the Simionescu shown in the main paper. Same as the main paper, in each figure, the true function landscape is on the left, where darker color means a higher objective value and the white region means infeasible. On the right is the predicted function landscape (top: CEI, middle: SCBO, bottom: BE-CBO) where darker color means a higher objective value and the contour means the feasibility boundary (feasible inside, infeasible outside). Initial samples (grey), the rest evaluated samples (black) and the global optima (red) are also displayed. 

For the LSQ function, the objective function is relatively smooth so all three algorithms discover the global optimum in the end. BE-CBO discovers the global optimum more efficiently at 100 evaluations and also classifies a much more accurate constraint boundary compared to the other two algorithms. For the Townsend function, both CEI and SCBO get stuck in the two local optima in the middle, while BE-CBO sucessfully discovers the true global optimum on the upper right corner and also classifies the constraint boundary well.

\subsection{Runtime Comparison}
\label{app:comp:runtime}

\begin{figure}[h]
    \centering
    \includegraphics[width=1.0\textwidth]{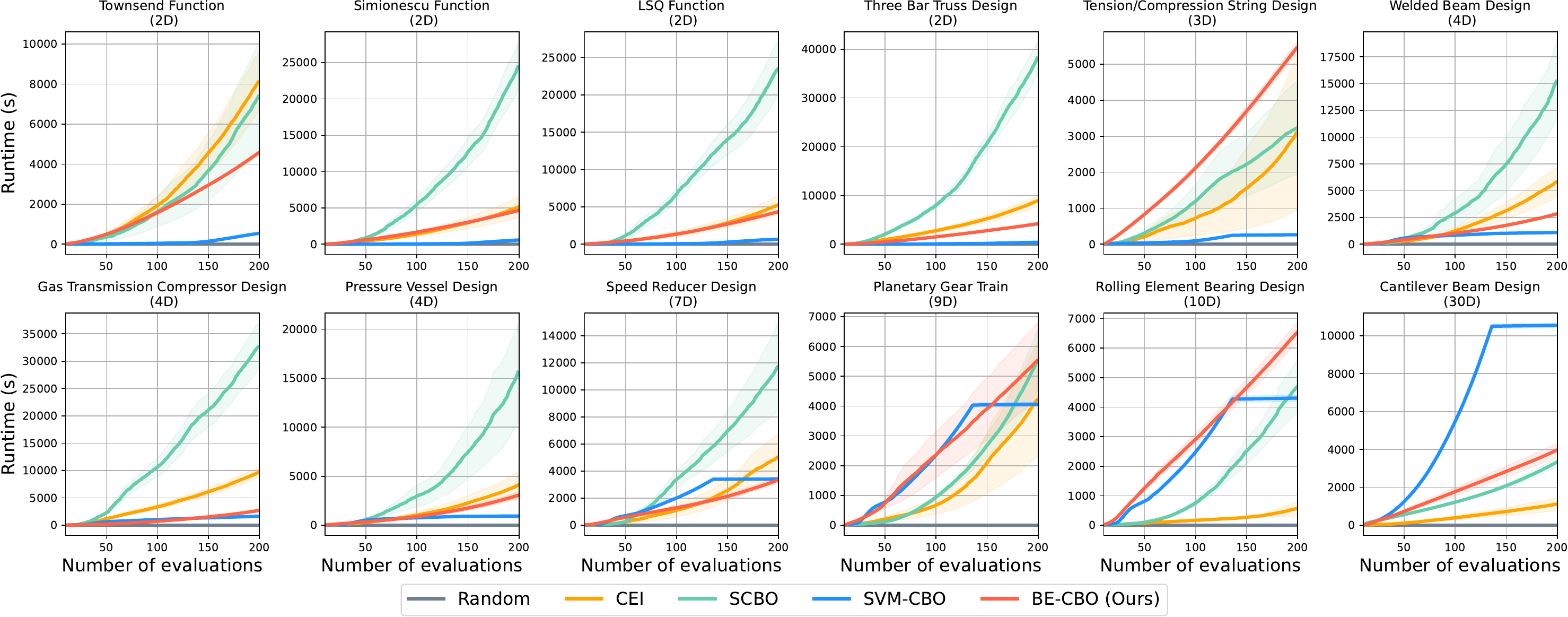}
    \caption{Runtime comparison between algorithms. The experiments are averaged over 10 random seeds. The time is accumulated after every iteration, i.e. after 200 evaluations the runtime is showing the total number of seconds spent to propose and evaluated all 200 samples.}
    \label{fig:runtime}
\end{figure}

In order to empirically assess the speed efficiency of various algorithms, we collect and analyze the runtime statistics measured in seconds, shown in Figure~\ref{fig:runtime}. 
In general, BE-CBO exhibits stable runtime across different problem dimensions with very low variance, at around 5K seconds for 200 iterations. Our speed is close to CEI in many cases and faster than SCBO on average.
One might observe that BE-CBO has slower performance than all other algorithms on Tension/Compression String Design, Planetary Gear Train and Rolling Element Bearing Design problems. Though in these cases BE-CBO's speed does not change much, other algorithms operate faster than on other problems. This observation can be actually explained by our proposed feasibility ratio metric shown in Figure~\ref{fig:feasibles}. In those cases, our algorithm is the most successful in discovering feasible points while other algorithms produce mostly infeasible points. The surrogate classifiers in other algorithms fail to learn useful information thus the fitting stops early, with the exception of SVM-CBO where the SVM fitting time scales badly with the problem dimension.

\subsection{Synthetic Benchmark Problems With Interior Optima}
\label{app:comp:interior}

\begin{figure}[!h]
    \centering
    \includegraphics[width=0.7\textwidth]{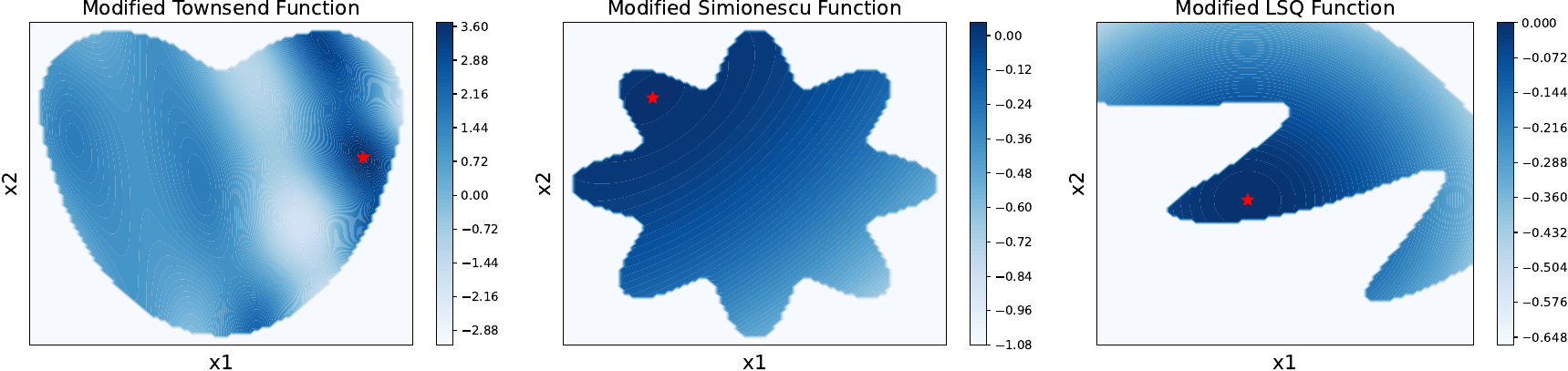}
    \caption{Landscape of the modified Townsend, Simionescu and LSQ functions. The red star indicates the global optimum location, which is inside the feasible region instead of on the constraint boundary in their original forms.}
    \label{fig:interior_fun}
\end{figure}

Although our motivation of developing BE-CBO is the observation that most real-world problems have their optima on the constraint boundary due to physical limits, to test the generality of our method, we construct synthetic functions with global optimum located at the interior of the feasible region. We modified the objective functions of Townsend, Simionescu and LSQ while leaving their constraint functions unchanged. Figure~\ref{fig:interior_fun} shows the landscape of the modified functions. The analytical forms of the modified objective functions are as follows:
\begin{itemize}
    \item Townsend: $f(x) = - [\cos(((x_1 + 1) - 0.1)x_2)]^2 - (x_1 + 1) \sin(3(x_1 + 1) + x_2)$
    \item Simionescu: $f(x) = 0.1x_1x_2 + 0.1(x_1 - x_2 + 1)^2$
    \item LSQ: $f(x) = (x_1 - 0.4)^2 + (x_2 - 0.45)^2$
\end{itemize}

\begin{figure}[!h]
    \centering
    \includegraphics[width=0.7\textwidth]{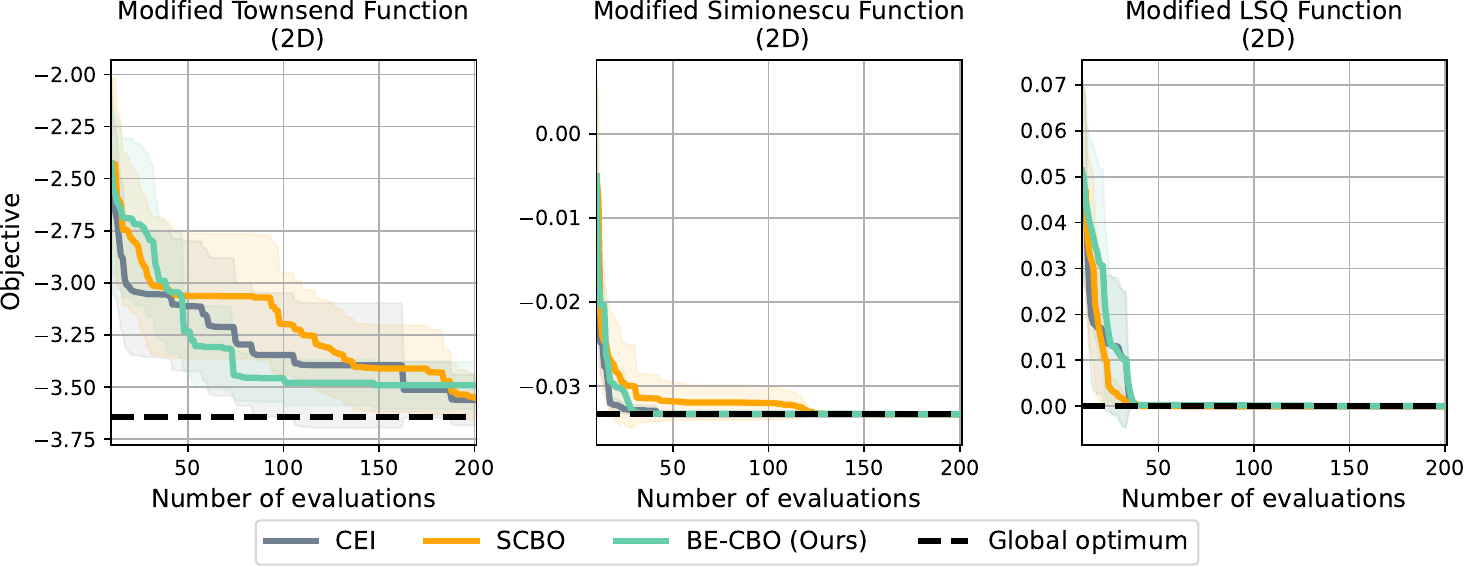}
    \caption{Quantitative comparison of different algorithms including our BE-CBO on shifted synthetic test functions. The current best value is shown w.r.t. the number of function evaluations. Every experiment has 10 initial random samples and 200 evaluations in total. The curve is averaged over 10 different initial random seeds and the standard deviation is shown as a shaded region.}
    \label{fig:interior_perf}
\end{figure}

We conduct experiments on these modified functions using CEI, SCBO and BE-CBO. The comparison results are shown in Figure~\ref{fig:interior_perf}. The results show that BE-CBO can effectively discover interior optima in different functions. Though BE-CBO encourages exploration on the constraint boundary, we leverage the underlying Expected Improvement acquisition to decide where is the most promising region considering both the dynamic band and the whole predicted feasible region. In other words, whether to explore the boundary region or the interior feasible region depends on the predicted objective landscape, thus both boundary and interior optima can be discovered.

\section{Ablation Studies}
\label{app:ablation}

\subsection{Gaussian Processes vs Deep Ensembles for Modeling Unknown Constraints}
\label{app:ablation:gp_de}

We use a surrogate model to approximate the unknown constraints. This surrogate model needs to work as a binary classifier that predicts whether a point is feasible since we do not get any additional information when the point is infeasible. Instead of using a popular choice for surrogate models in Bayesian optimization, which are Gaussian Processes, here we show how Deep Ensembles can benefit the overall performance of our Bayesian optimization framework.

\subsubsection{Classification Performance Comparison}
\label{app:ablation:gp_de:cls_acc}

We demonstrate how the classification accuracy of Deep Ensembles is more stable and in general outperforms the accuracy of Gaussian Processes (GPs). In real-world problems, it is often hard to find any feasible points due to complex constraints, which is reflected in some of the benchmark problems we used to test our approach (see further details in Section~\ref{app:setup:problems}). In such cases, we will typically obtain imbalanced data where most of the points are infeasible. Hence, to properly compare the classification accuracy, we use the \textit{Balanced Accuracy}~\citep{kelleher2020fundamentals} metric, defined as \[\frac{TPR+TNR}{2}\] where $TPR=\frac{\text{true positive}}{\text{total positive}}$ is a true positive rate measuring the sensitivity and $TNR=\frac{\text{true negative}}{\text{total negative}}$ is a true negative rate measuring the specificity. 

\begin{figure}[!h]
    \centering
    \includegraphics[width=1.0\textwidth]{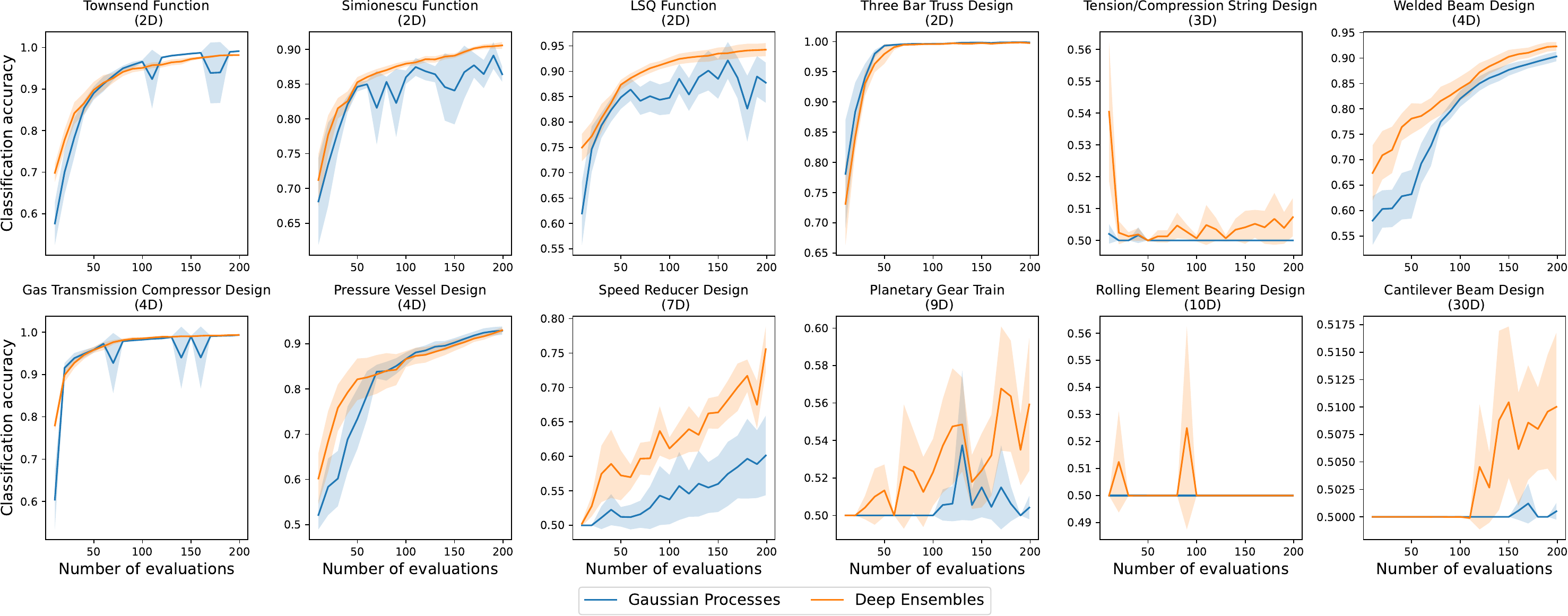}
    \caption{Comparison of classification accuracy between Gaussian Processes and Deep Ensembles when used as classifiers in our Bayesian optimization framework. Classification accuracy is computed with the \textit{Balanced Accuracy} metric. Curves show the average over 10 random seeds and shaded regions represent standard deviation. }
    \label{fig:classif_acc}
\end{figure}

The experiments are conducted over the same 10 random seeds for both Deep Ensembles and GPs. To test the GPs we simply replace the Deep Ensembles with GPs in our algorithm BE-CBO to train the classifier which models the constraints. All the other steps of our Bayesian optimization framework are identical. Starting from 10 initial random samples, we perform 200 function evaluations in each test run and check the classification accuracy after each evaluation when the classifiers are updated. The results are reported in Figure~\ref{fig:classif_acc}. Note that GPs oscillate in accuracy between different iterations and perform poorly on complex real-world problems, which immediately affects the overall algorithm performance shown in Figure~\ref{fig:gpens_perf}.   

\subsubsection{Effect of Gaussian Processes vs Deep Ensembles on BE-CBO}
\label{app:ablation:gp_de:ours}
\begin{figure}[!h]
    \centering
    \includegraphics[width=1.0\textwidth]{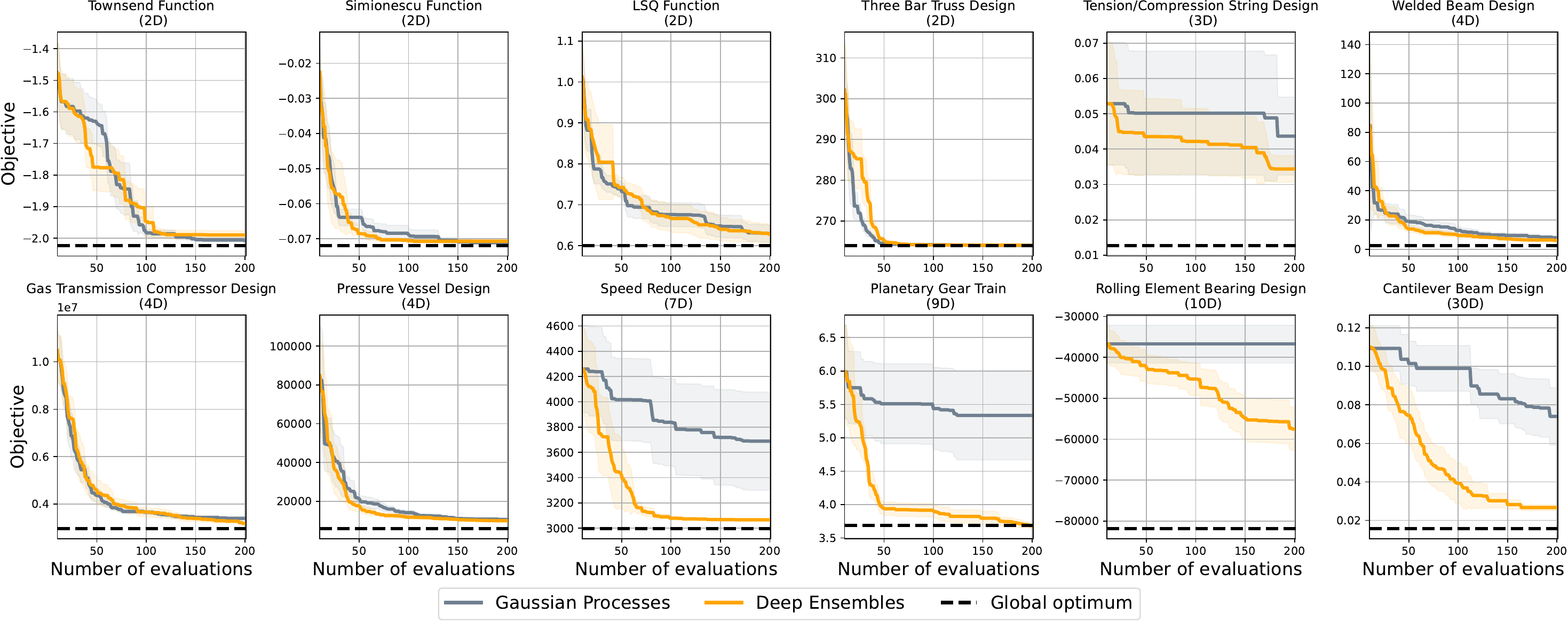}
    \caption{Performance of our algorithm BE-CBO when using Deep Ensembles vs Gaussian Processes for modeling unknown constraints. The current best value found by an algorithm is shown w.r.t. the number of function evaluations. Experiments are run independently from 10 random seeds. The bold curves represent the average in performance over all 10 seeds and shaded areas reflect the standard deviation.}
    \label{fig:gpens_perf}
\end{figure}

We compare the performance of our algorithm BE-CBO when using different surrogate models for the unknown constraints. In one setup, we run our proposed algorithm consisting of Deep Ensembles for modeling the constraints, while in the other setup, we replace the Deep Ensembles with Gaussian Processes (GPs). Both setups are tested on the same 10 initial random samples and we report the average performance and the standard deviation from 10 random seeds in Figure~\ref{fig:gpens_perf}. Deep Ensembles and GPs have comparable performance for simpler benchmark problems, while Deep Ensembles demonstrate superior performance in higher dimensions and Tension/Compression String design that is known to be challenging to model and find any feasible designs. Furthermore, we note that GPs perform well only when they are able to closely approximate the true boundary between the feasible and infeasible region of the design space. This finding is reflected in the classification accuracy of different surrogate models, as seen in Section~\ref{app:ablation:gp_de:cls_acc}. 

\subsection{Hyperparameters of Deep Ensembles}
\label{app:ablation:de}

\subsubsection{Number of MLPs}
\label{app:ablation:de:n_mlp}
\begin{figure}[!h]
    \centering
    \includegraphics[width=1.0\textwidth]{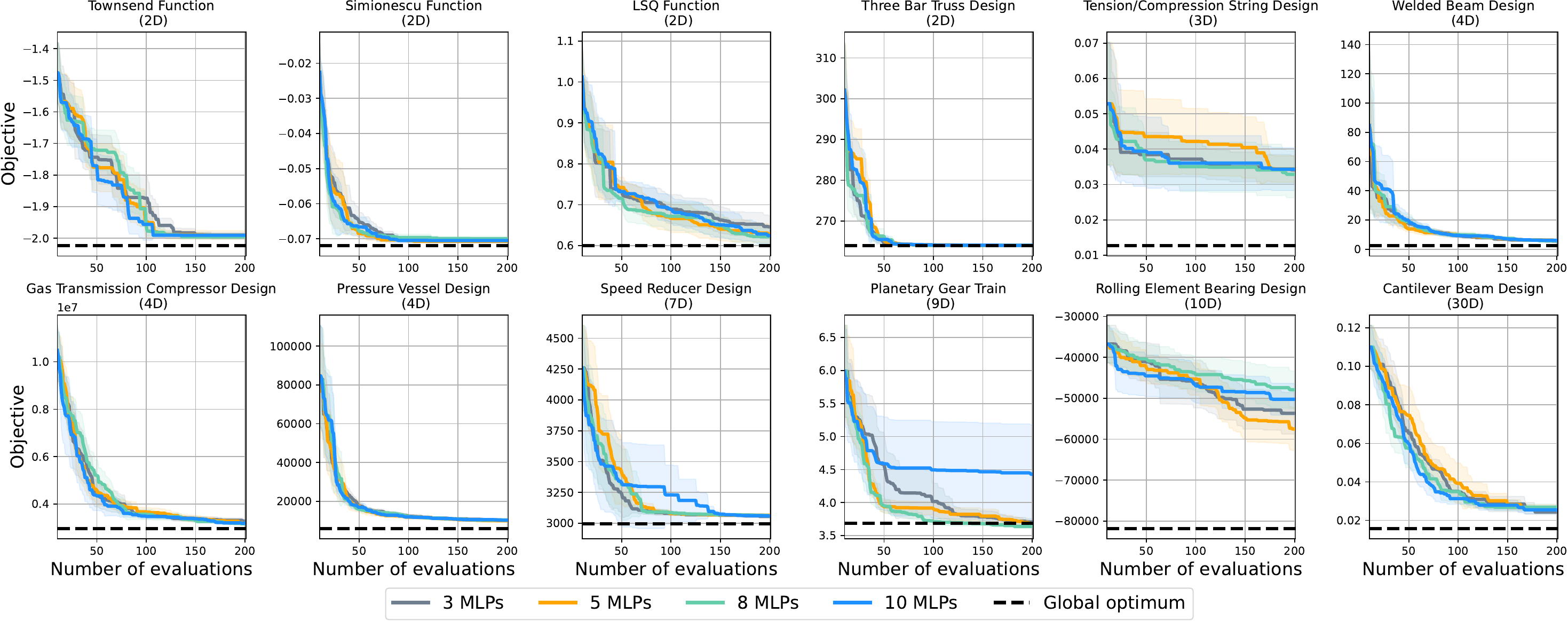}
    \caption{Performance comparison of our algorithm when using different number of MLPs in ensemble. The current best value found by an algorithm is shown w.r.t. the
number of function evaluations and the performance is averaged over 10 initial seeds for each experiment. }
    \label{fig:n_mlps}
\end{figure}

One hyperparameter in our network ensemble is the number of MLPs used. However, we observed that varying this parameter does not have much significant effect on the overall performance of our algorithm. In our experiments, we use 5 MLPs in an ensemble for efficiency in memory and computing resources. 
Our tests show that increasing the number of MLPs up to 8 or decreasing down to 3 mostly has little effect on the overall performance (see Figure~\ref{fig:n_mlps}).

\subsubsection{Number of Hidden Layers}
\label{app:ablation:de:n_layers}

\begin{figure}[!h]
    \centering
    \includegraphics[width=1.0\textwidth]{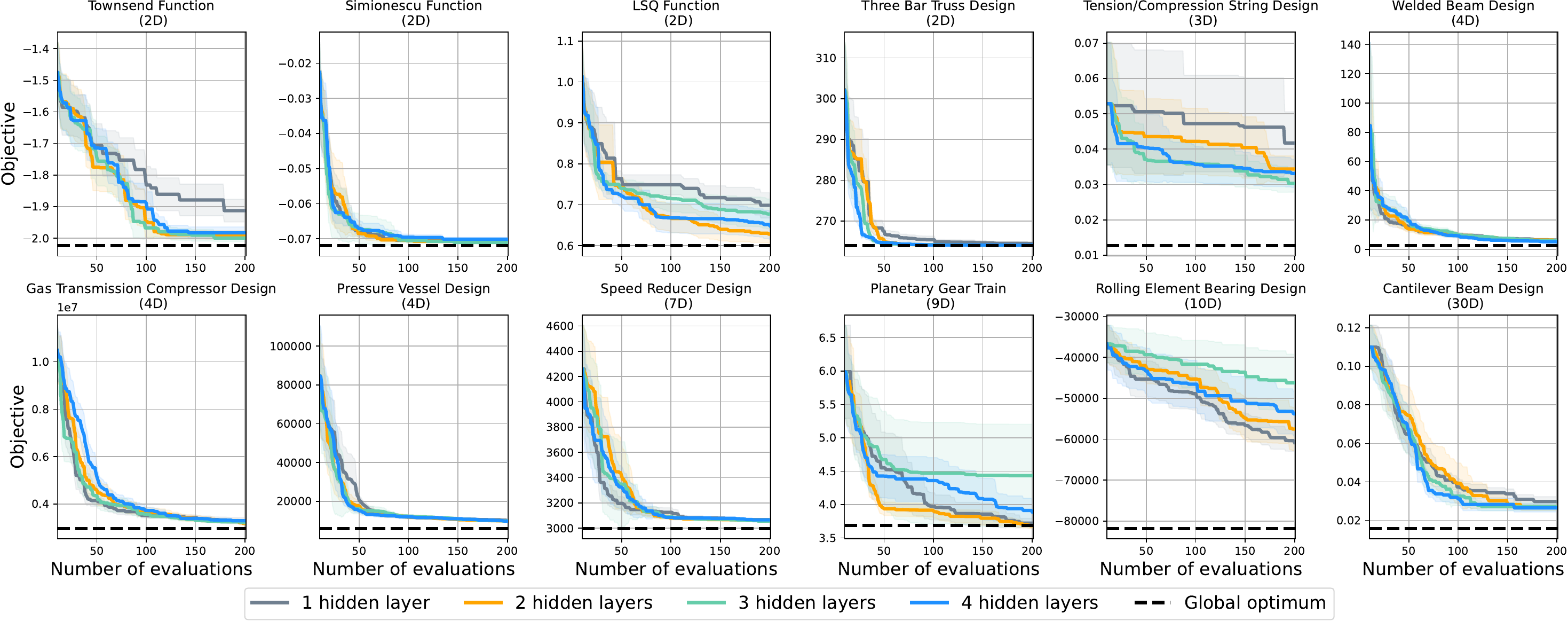}
    \caption{Comparison of BE-CBO with different number of hidden layers for the Deep Ensemble classifier. The current best value is shown w.r.t. the number of function evaluations. The curve is averaged over 10 different initial random seeds. The performance of BE-CBO is not particularly sensitive to the number of layers as long as it has more than one hidden layer.}
    \label{fig:n_hidden_layers}
\end{figure}

The effect of the number of layers on the algorithm is shown in Figure~\ref{fig:n_hidden_layers}. We tested 1, 2, 3, 4 hidden layers with $64 \log_2{d}$ neurons in each layer where $d$ is the problem dimension. The result shows that one hidden layer is insufficient in some cases since the network is too shallow which leads to underfitting. Two or more hidden layers perform similarly in general.

\subsubsection{Number of Neurons in a Layer}
\label{app:ablation:de:n_neurons}

\begin{figure}[!h]
    \centering
    \includegraphics[width=1.0\textwidth]{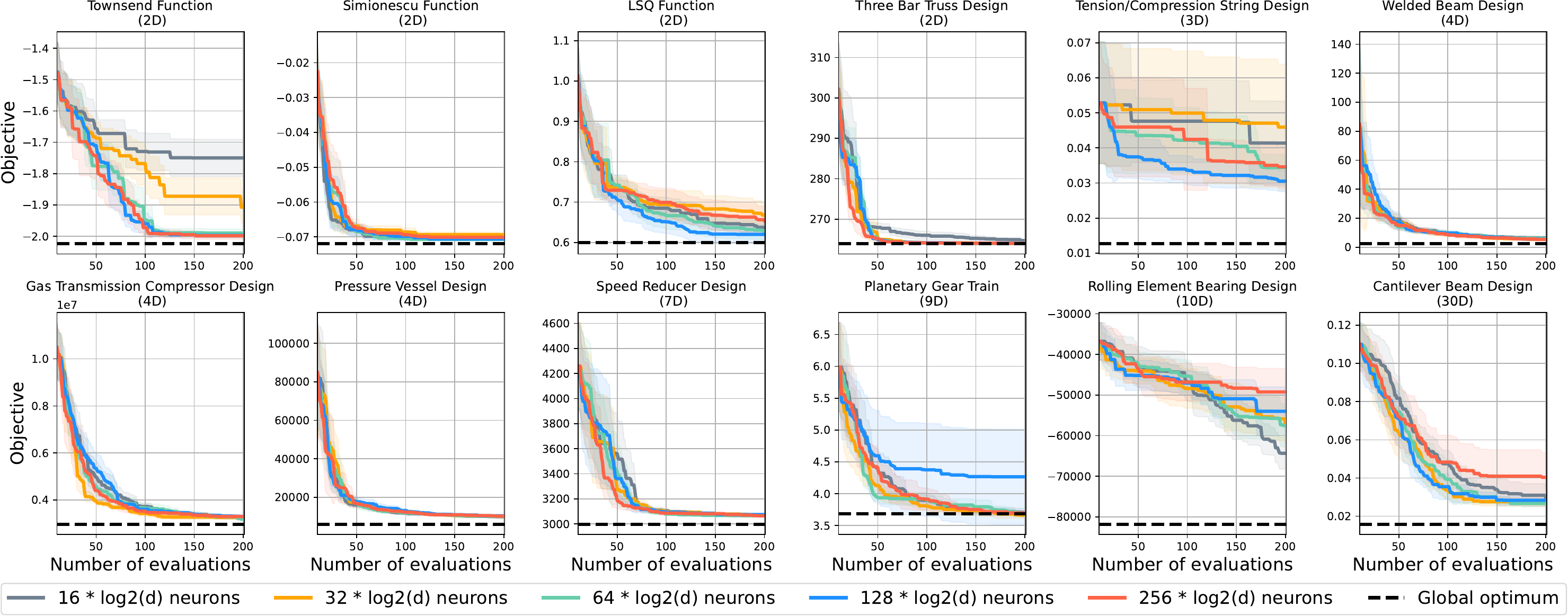}
    \caption{Comparison of BE-CBO with different number of neurons in a layer for the Deep Ensemble classifier. The current best value is shown w.r.t. the number of function evaluations. The curve is averaged over 10 different initial random seeds. The performance of BE-CBO is not sensitive to the number of neurons as long as it has at least 64 neurons in a layer.}
    \label{fig:n_neurons}
\end{figure}

The effect of the number of layers on the algorithm is shown in Figure~\ref{fig:n_neurons}. To effectively learn across all problem dimensions, we scale the number of neurons in each layer w.r.t. the problem dimension following a logarithmic formula $N(d) = C \log_2(d)$ where $N$ is the number of neurons, $d$ is the problem dimension and $C$ is a constant factor. We tested $C = 16, 32, 64, 128, 256$ with 2 hidden layers. The results suggest that on most problems, the number of neurons do not matter much, but in some problems, networks with a low capacity ($C =$ 16 or 32) are outperformed by networks with more neurons and it seems like $C = 64$ is a stable choice across all problems.

\subsubsection{Learning Rate}
\label{app:ablation:de:lr}

\begin{figure}[!h]
    \centering
    \includegraphics[width=1.0\textwidth]{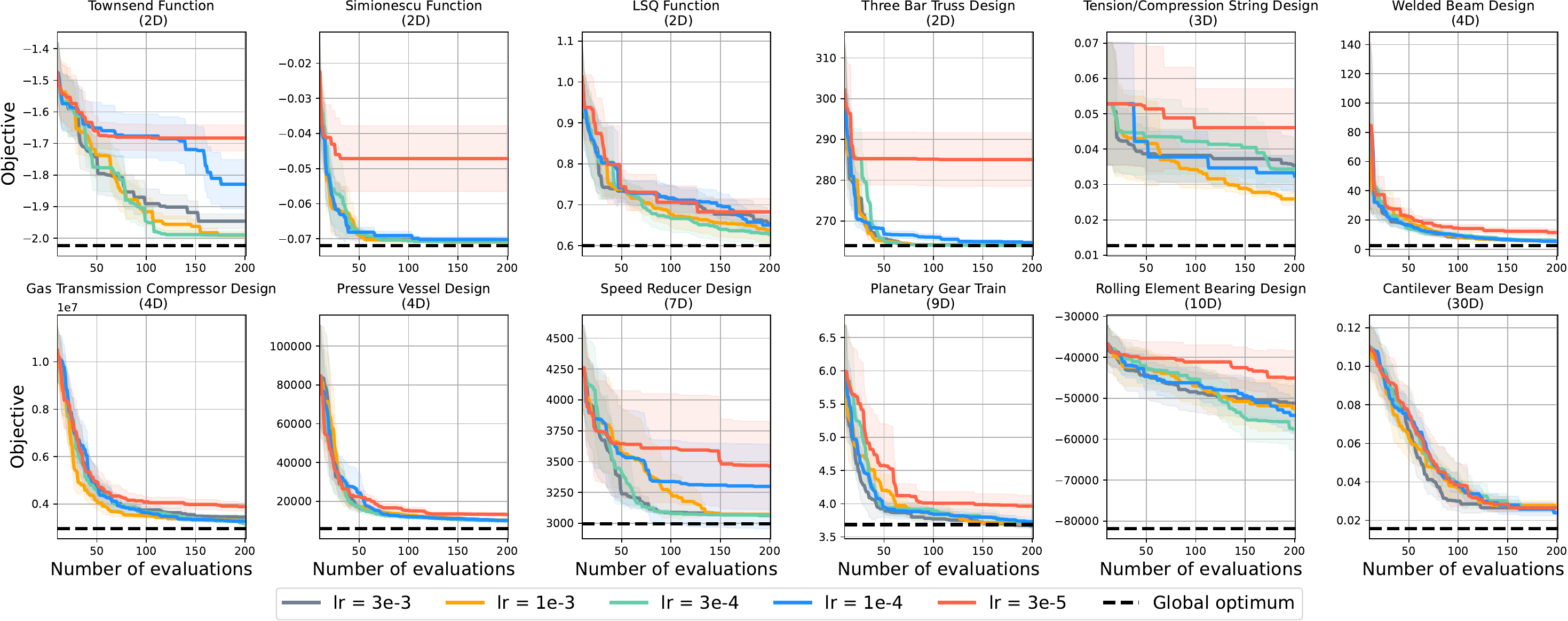}
    \caption{Comparison of BE-CBO with different learning rates for training the Deep Ensemble classifier. The current best value is shown w.r.t. the number of function evaluations. The curve is averaged over 10 different random seeds.
    Relatively large learning rates perform similarly good as small rates will underfit the network since the number of training iterations is fixed.}
    \label{fig:lr}
\end{figure}

In the paper, we did experiments with a 3e-4 learning rate, which is a common choice for this hyperparameter. We now test 3e-3, 1e-3, 3e-4, 1e-4, 3e-5 learning rates and show results in Figure~\ref{fig:lr}. Since we set a fixed number of training iterations (1000), small learning rates 1e-4 and 3e-5 are outperformed by larger rates because the network training does not converge. Larger learning rates (3e-3, 1e-3, 3e-4) perform similarly with slight variations on different problems.

\subsection{Acquisition Function Choices}
\label{app:ablation:acq}

\begin{figure}[!h]
    \centering
    \includegraphics[width=1.0\textwidth]{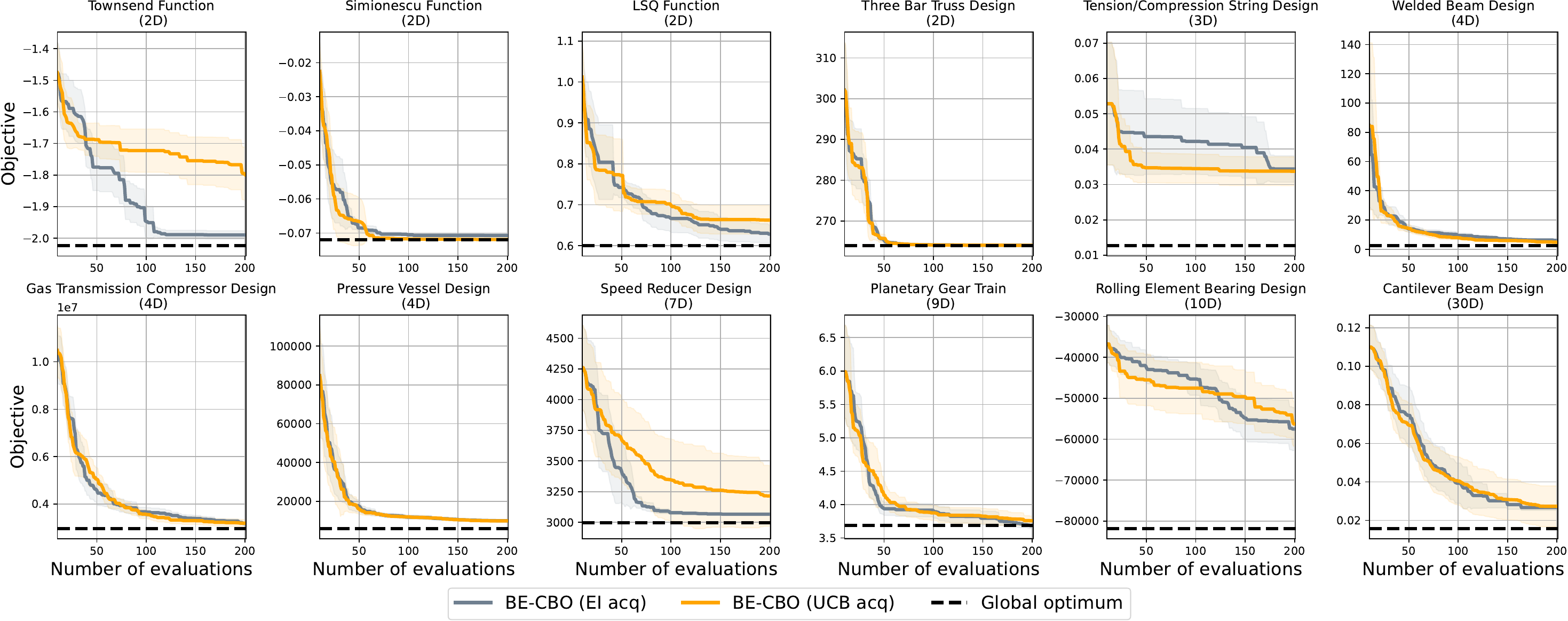}
    \caption{Comparison of BE-CBO with different acquisition functions (EI and UCB). The current best value is shown w.r.t. the number of function evaluations. The curve is averaged over 10 different random seeds.}
    \label{fig:ei_ucb}
\end{figure}

To understand the performance of BE-CBO on more acquisition functions, we switch from EI to UCB (with beta = 0.1 as the default value in BoTorch) in our method and run empirical comparisons. As shown in Figure~\ref{fig:ei_ucb}, the results suggest that EI and UCB perform similarly well on our benchmark problems overall, with some performance differences in particular problems. In practice, users may select the proper acquisition function to be used in BE-CBO based on the desired properties (such as the explicit control over exploration in UCB).

\subsection{Dynamic Adaptation of Bounds for Boundary Exploration}
\label{app:ablation:be}

To validate whether the proposed boundary exploration strategy is successful, we modify our BE-CBO with different methods of coupling the constraint into the acquisition function instead of doing boundary exploration. For example, CEI proposes to multiply the feasible probability with the acquisition function, and SCBO applies a $0.5$ upper bound on the feasible probability as a constraint on top of the acquisition function. Specifically, when ignoring the rest of their approaches, they can be written as optimizing:
\[\text{CEI:} \quad \argmax_x{C(x)q(x)} \]
\[\text{SCBO:} \quad \argmax_x{q(x)} \quad \text{s.t.} \quad C(x) \geq 0.5 \]
For reference, our approach is optimizing:
\[\text{BE-CBO:} \quad \argmax_x{q(x)} \quad \text{s.t.} \quad C(x) \geq 0.5 - \sigma_E(x) \]

\begin{figure}[!h]
    \centering
    \includegraphics[width=1.0\textwidth]{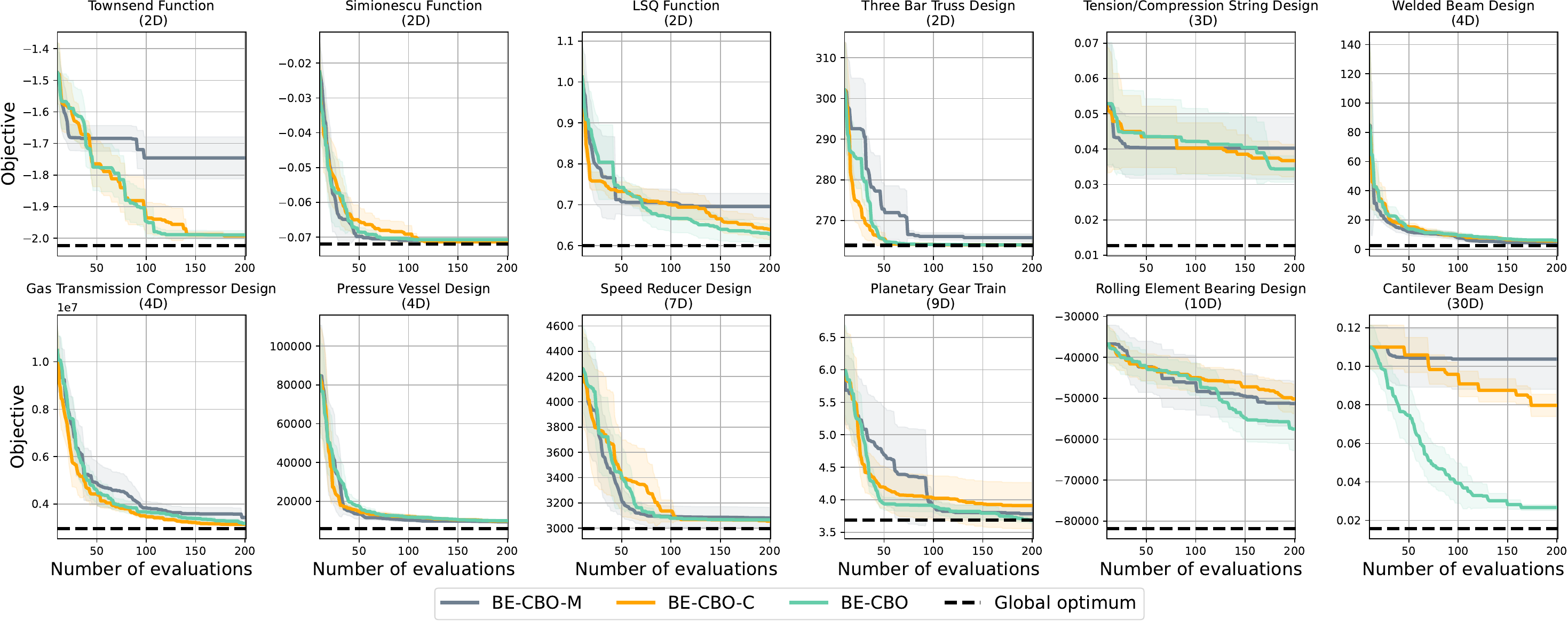}
    \caption{Comparison of BE-CBO with different variants of constraint coupling in the acquisition function. The current best value is shown w.r.t. the number of function evaluations. The curve is averaged over 10 different initial random seeds. }
    \label{fig:ablation_be}
\end{figure}

Thus, we develop two variants of BE-CBO based on CEI and SCBO's optimization objective and call them BE-CBO-M (M stands for multiplication of $C(x)$) and BE-CBO-C (C for the $0.5$ cutoff on $C(x)$), while the rest of the algorithm (e.g., surrogate models) share the same design choices as BE-CBO. Figure~\ref{fig:ablation_be} compares the performance between these two variants and BE-CBO on all benchmark problems. 

The results show that BE-CBO-M is clearly outperformed by BE-CBO-C and BE-CBO. Even though the multiplicative form of BE-CBO-M is a reasonable design in theory, in practice, the inaccuracy in constraint modeling can lead to failure cases easily. On one hand, multiplying $C(x)$ with the acquisition may encourage sampling in the safe region instead of exploration; On the other hand, when overestimation of the acquisition value happens in the infeasible region, even with a small probability of $C(x)$, it may still sample very far away in the infeasible region where the predicted $q(x)$ is huge. The algorithm gets stuck in such scenario because the new evaluated infeasible point does not update the classifier much since it already has a low feasibility prediction, then in the next iteration, the algorithm will keep proposing points in the similar region.

\begin{figure}[!ht]
    \centering
    \includegraphics[width=0.8\textwidth]{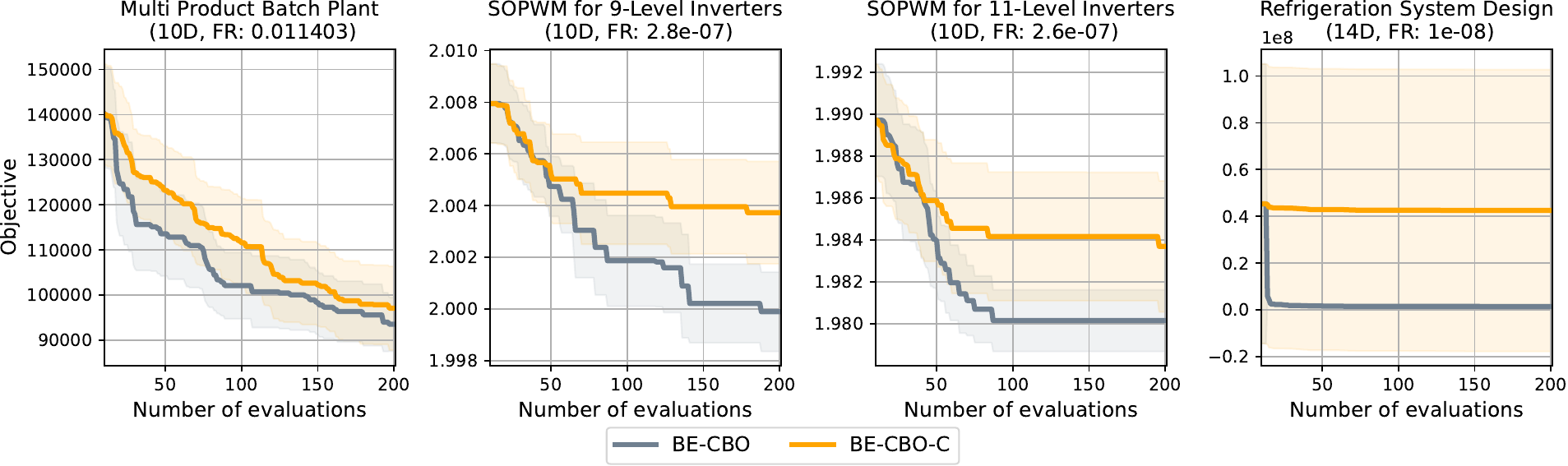}
    \caption{Comparison between BE-CBO and BE-CBO-C on additional high-dimensional real-world problems.}
    \label{fig:ablation_be_highdim}
\end{figure}

\begin{figure}[!h]
    \centering
    \includegraphics[width=1.0\textwidth]{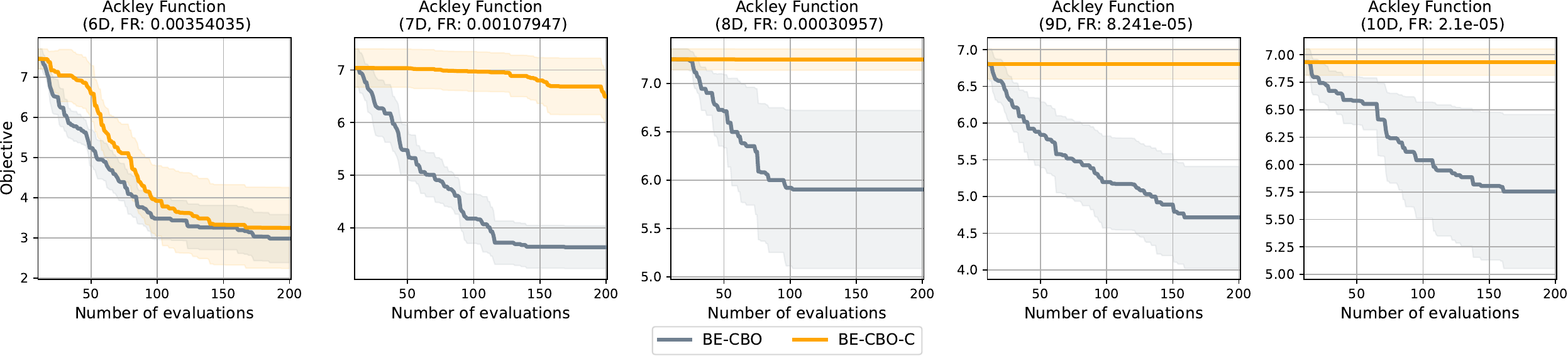}
    \caption{Comparison between BE-CBO and BE-CBO-C on additional synthetic Ackley functions with different dimensionalities.}
    \label{fig:ablation_be_ack}
\end{figure}

BE-CBO-C shares a similar performance with BE-CBO in low dimensional problems, but the performance starts to deteriorate when evaluated on high dimensional problems. Especially on the 30D Cantilever Beam Design problem, having a $0.5$ cutoff bound in the objective makes BE-CBO-C explores much slower compared to BE-CBO. For more thorough evaluations, we implemented four additional high-dimensional real-world benchmark problems, including 10D multi-product batch plant design~\cite{grossmann1979optimum}, 10D synchronous optimal pulsewidth modulation control for 9-level and 11-level inverters~\cite{rathore2010synchronous}, and 14D industrial refrigeration system design~\cite{paul1987optimal}. Besides the dimensionality, we would like to highlight the feasibility ratio (FR) as an important property of our benchmark problems, which indicates the ratio of feasible samples among a massive amount of random samples in the parameter space. As shown in Figure~\ref{fig:ablation_be_highdim}, BE-CBO greatly outperforms BE-CBO-C on high-dimensional problems with a low feasibility ratio.
To further validate this observation, we conduct controlled experiments on standard synthetic Ackley functions (constrained in the same way following SCBO) with varying dimensions. Figure~\ref{fig:ablation_be_ack} shows that BE-CBO becomes more advantageous than BE-CBO-C on problems with higher dimensions and lower feasibility ratios, and BE-CBO-C completely fails to improve in challenging cases. In such scenarios, exploration, or the ability to jump out of local minima, is crucial to the algorithm’s performance. BE-CBO promotes exploration through our proposed dynamic bound strategy, while the fixed and conservative bound of BE-CBO-C discourages it from effective exploration.

\end{document}